\documentclass{sn-jnl}

\usepackage[numbers]{natbib}   
\usepackage{multirow,colortbl,tablefootnote,graphicx,epstopdf,color,booktabs,amsmath,amssymb,diagbox}
\usepackage{titlesec} 
\usepackage{balance}
\usepackage{hyperref}
\hypersetup{
    colorlinks=true,
    linkcolor={black},
    citecolor={blue},
    urlcolor={black}
}
\usepackage{xspace}
\usepackage{enumitem}
\usepackage[table]{xcolor}
\usepackage[font=footnotesize]{subcaption}

\usepackage[noabbrev,nameinlink]{cleveref}
\crefname{property}{property}{Property}
\creflabelformat{property}{(#1)#2#3}
\crefname{equation}{eq}{Eq}
\creflabelformat{equation}{(#1)#2#3}

\newtheorem{definition}{Definition}

\newlength{\bibitemsep}
\newlength{\bibparskip}
\let\oldthebibliography\thebibliography
\renewcommand\thebibliography[1]{%
  \oldthebibliography{#1}%
  \setlength{\parskip}{\bibitemsep}%
  \setlength{\itemsep}{\bibparskip}%
}

\newcommand\blfootnote[1]{%
  \begingroup
  \renewcommand\thefootnote{}\footnote{#1}%
  \addtocounter{footnote}{-1}%
  \endgroup
}

\newenvironment{tbox}{\begin{tcolorbox}[
		enlarge top by=5pt,
		enlarge bottom by=5pt,
		 breakable,
		 boxsep=0pt,
                  left=4pt,
                  right=4pt,
                  top=10pt,
                  arc=0pt,
                  boxrule=1pt,toprule=1pt,
                  colback=white
                  ]
	}
{\end{tcolorbox}}

\makeatletter
\newcommand{\thickhline}{%
    \noalign {\ifnum 0=`}\fi \hrule height 1.3pt
    \futurelet \reserved@a \@xhline
}

\newcommand{\prob}[1]{\Pr\left[#1\right]}

\begin{document}

\title{Deconstructing The Ethics of Large Language Models from Long-standing Issues to New-emerging Dilemmas: A Survey}

\author{Chengyuan Deng$^{1}\dagger$, Yiqun Duan$^{2}\dagger$, Xin Jin$^{3}\dagger$, Heng Chang$\dagger$, Yijun Tian$^{5}$, Han Liu$^{6}$, Yichen Wang$^{7}$, Kuofeng Gao, Henry Peng Zou$^{8}$, Yiqiao Jin$^{9}$, Yijia Xiao$^{10}$,  Shenghao Wu$^{11}$, Zongxing Xie$^{12}$, Weimin Lyu$^{12}$, Sihong He$^{13}$, Lu Cheng$^{8}$, Haohan Wang$^{14}$, Jun Zhuang$^{15}$\\
 $^{1}${Rutgers University},
 $^{2}${University of Technology Sydney},
 $^{3}${Ohio State University},
 $^{5}${University of Notre Dame},
 $^{6}${Washington University in St. Louis},
 $^{7}${University of Chicago},
 $^{8}${University of Illinois Chicago},
 $^{9}${Georgia Institute of Technology},
 $^{10}${University of California, Los Angeles},
 $^{11}${Carnegie Mellon University},
 $^{12}${Stony Brook University},
 $^{13}${University of Connecticut},
 $^{14}${University of Illinois Urbana-Champaign},
 $^{15}${Boise State University}\\
\vspace{-1cm}
\email{$^{15}$junzhuang@boisestate.edu}
}

\abstract{Large Language Models (LLMs) have achieved unparalleled success across diverse language modeling tasks in recent years. However, this progress has also intensified ethical concerns, impacting the deployment of LLMs in everyday contexts. This paper provides a comprehensive survey of ethical challenges associated with LLMs, from longstanding issues such as copyright infringement, systematic bias, and data privacy, to emerging problems like truthfulness and social norms. We critically analyze existing research aimed at understanding, examining, and mitigating these ethical risks. Our survey underscores integrating ethical standards and societal values into the development of LLMs, thereby guiding the development of responsible and ethically aligned language models.}


\maketitle

\section{Introduction}
\label{sec:introduction}
\blfootnote{$\dagger$ All authors contributed equally.}

In the past few years, the field of artificial intelligence (AI) has witnessed a surge in the development of large language models (LLMs). These advanced computational language models have demonstrated remarkable performance across a spectrum of language modeling tasks~\cite{chang2023survey, su2024large, wang2023lemon, zhao2023competeai, zhuang2024understanding, zhuang2024robust, mao2024compressibility, yang2024editworld}. Their capabilities are exemplified in natural language generation~\cite{brown2020language, chen2023semi, mo2024large}, where they can create coherent and contextually relevant text, question answering~\cite{baek2023knowledge, zhang2024unlocking, zou2024eiven}, where they effectively retrieve or infer information in response to queries, and complex reasoning tasks~\cite{huang2022towards, jin2023binary, zhang2024high, yan2023comprehensive}, which involve navigating through intricate problem-solving processes. Despite these advancements, LLMs have also raised substantial ethical concerns. As these models become increasingly integrated into daily life, addressing these ethical challenges becomes paramount. The concerns are multifaceted, encompassing issues such as privacy~\cite{xiao2023large}, copyright, robustness~\cite{zhang2023foundation}, bias, and the potential for misuse. Given their ability to understand and generate human-like responses, there's a growing discourse on ensuring these responses are not only accurate but also ethically aligned with societal norms and values.

In response to ethical concerns, substantial research is focusing on the ethical implications of LLMs. Scholars aim to identify, examine, and mitigate potential risks, guiding the development of more responsible AI systems \cite{cheng2021socially}. This effort ensures LLMs are designed and deployed to maximize benefits and minimize harm, serving the public good ethically and effectively. The realization of these objectives hinges heavily on access to large-scale high-quality corpus and textual datasets. However, collecting the data may bring ethical issues, such as privacy, copyright, and bias~\cite{xiao2023large}. These ethical issues are long-existing and still challenging.
Besides, some new ethical issues emerge as LLMs develop. For example, there is a growing concern over the potential for LLMs to produce inappropriate responses to unethical queries. To avoid this issue, alignment techniques are developed to align the answers with human values~\cite{liu2023trustworthy}. Similarly, the phenomenon of model-generated content that lacks factual basis, often referred to as ``hallucinations'', presents another ethical concern~\cite{zhang2023siren}. Furthermore, some new issues may emerge during the applications of LLMs, such as law and regulatory compliance~\cite{lai2023large}.
To illustrate, we outline the significant ethical issues for each subsection as follows:
\begin{itemize}[leftmargin=0.12in]
    \item {\bf Privacy} issues brought by LLMs include but are not limited to memorization (or data leaking), and privacy attacks. To provide a comprehensive review of ethics issues in privacy concerns, we first introduce existing privacy issues and their challenges and further provide two aspects of alleviating methods, differentiable privacy LLMs and emerging methods of preserving privacy.
    \item {\bf Copyright} concerns may be raised in LLM-generated content. We chronologically introduce two main technology arms of copyright - backdoor and watermark - to demonstrate their expansion and diffusion. For example, our introduction ranges from protecting web texts by HTML coding to preserving general texts on embodied watermarks, and from protecting the outputs to safeguarding the generative model and datasets, etc.
    \item {\bf Fairness} problems, such as societal biases in the training data of LLMs, may cause harm to marginalized communities, like prejudices, stereotypes, and discriminatory attitudes. To provide inclusive and equitable LLM-based services, it is critical to prevent LLMs from unintentionally perpetuating or amplifying these biases when generating responses.
    \item {\bf Truthfulness} of LLMs may be undermined by hallucination and sycophancy issues. Specifically, hallucination problems may inadvertently result in generating false information that appears credible, whereas sycophancy issues may amplify human preference rather than correct response. Addressing these two concerns is crucial to maintaining the trust and credibility of LLM technologies.
    \item {\bf Social Norm} plays a pivotal role in our society. However, LLMs may produce toxic content due to the contamination of train data. Alignment is one of the crucial techniques to address toxicity. In this survey, we discuss the motivation, characteristics, and recent advancements in alignment techniques, which are critical in the development and deployment of LLMs.
    \item {\bf Law and Regulatory Compliance} for LLMs are essential in our society as worldwide governments urgently promote AI-related legislation, such as the EU AI Act, to ensure that the utilization of AI tools aligns with ethical standards.
\end{itemize}

\begin{figure}[h]
\centering
\includegraphics[width=0.9\linewidth]{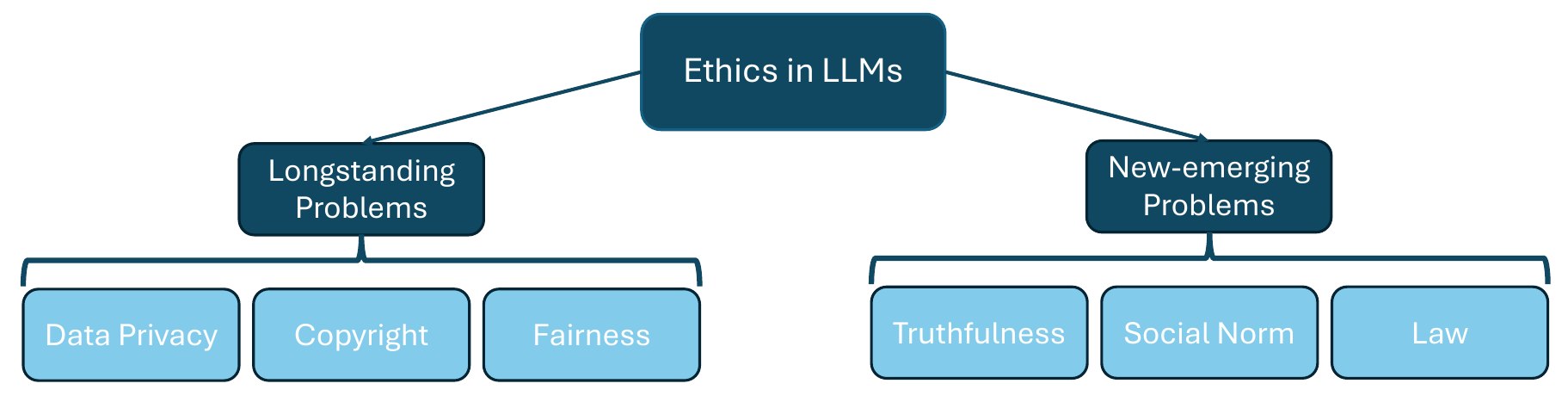}
\caption{Main category in this survey paper.}
\label{fig:category}
\end{figure}

In this survey, we aim to investigate ethical issues in the development of LLMs and propose a new taxonomy to help readers better understand the ethical issues and corresponding techniques that are proposed to solve these issues. Specifically, we categorize the ethical issues as longstanding problems and new-emerging problems. In the former category, we mainly discuss the ethical problems in 1) data privacy, 2) copyright, and 3) fairness. For the latter category, we are interested in the topic of truthfulness and social norms. Also, We introduce the law and regulatory compliance in the era of LLMs. To better illustrate our proposed taxonomy, we present the overall hierarchy in Figure~\ref{fig:category}.
In brief, we summarize our contributions in this survey as follows:
\begin{itemize}[leftmargin=0.12in]
    \item We systematically summarize and categorize existing ethical issues into two main categories: 1) we discuss {\bf longstanding} problems of data privacy, copyright, and fairness; 2) we investigate {\bf new-emerging} problems that are pertinent to LLMs, including truthfulness and social norms, and further discuss the design and requirement of law and regulatory compliance in guiding future explorations.
    \item We introduce the existing issues and mitigation strategies, and further present the hierarchy for each category in Figure~\ref{fig:long} and Figure~\ref{fig:new}.
    \item We discuss the future research directions for each section of the ethical issues.
\end{itemize}

The subsequent sections of this paper are structured as follows: Section~\{\ref{sec:lei}\} delves into enduring ethical dilemmas predating the advent of LLMs, while Section~\{\ref{sec:nei}\} introduces newly emergent ethical concerns in the era of LLMs.

\section{Persistent Ethical Issues}
\label{sec:lei}

\begin{figure}[h]
\centering
\includegraphics[width=\linewidth]{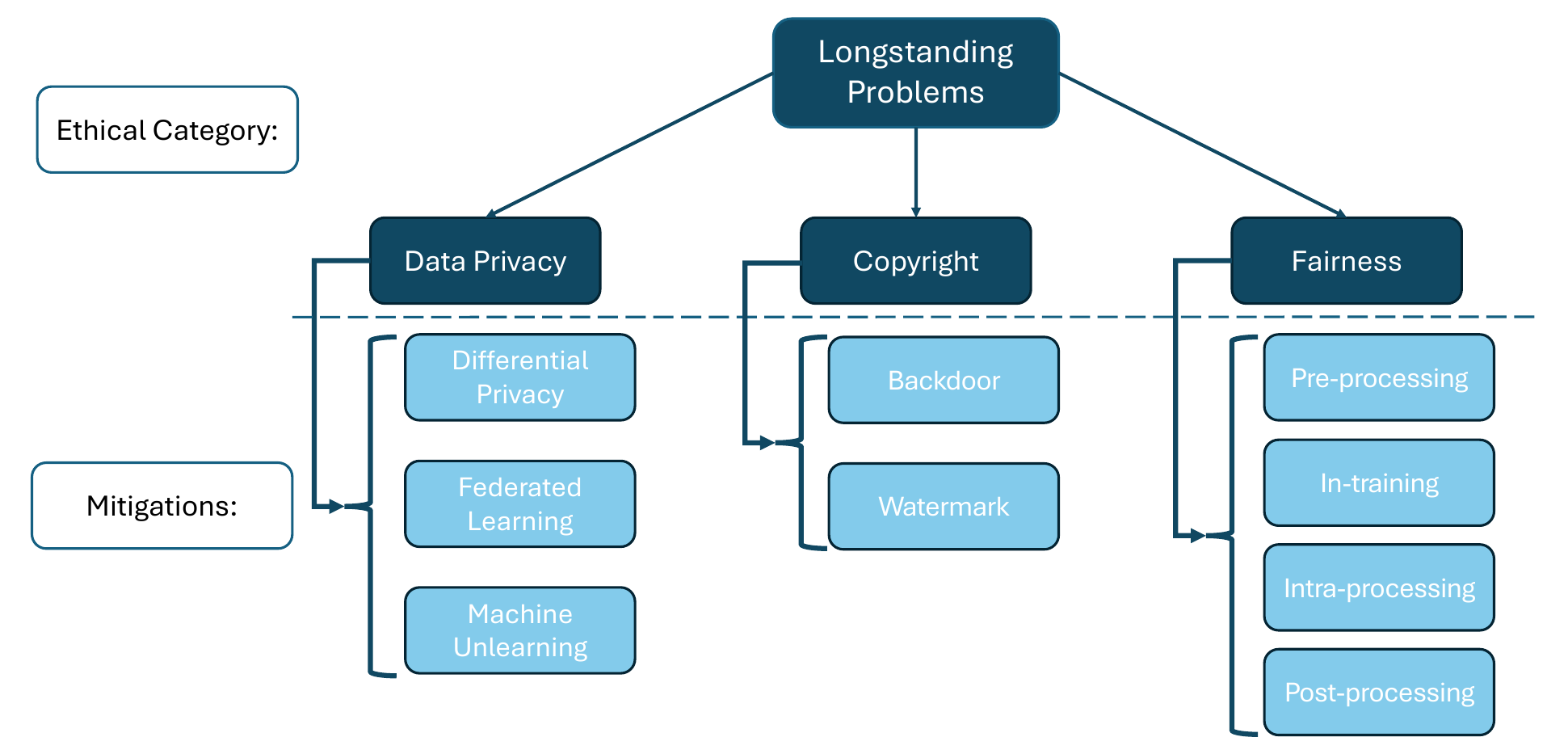}
\caption{The hierarchy of longstanding ethical problems in Section~\ref{sec:lei}. We list corresponding mitigation strategies for each sub-category.}
\label{fig:long}
\end{figure}

In this section, we present the longstanding ethical problems predating the advent of LLMs. These include 1) data privacy, 2) copyright, and 3) fairness. The hierarchy is displayed in Figure~\ref{fig:long}.

\subsection{Data Privacy}
\label{sec:privacy}

\subsubsection{Privacy: Issues and Challenges}
\begin{figure}[h]
\centering
\includegraphics[width=\linewidth]{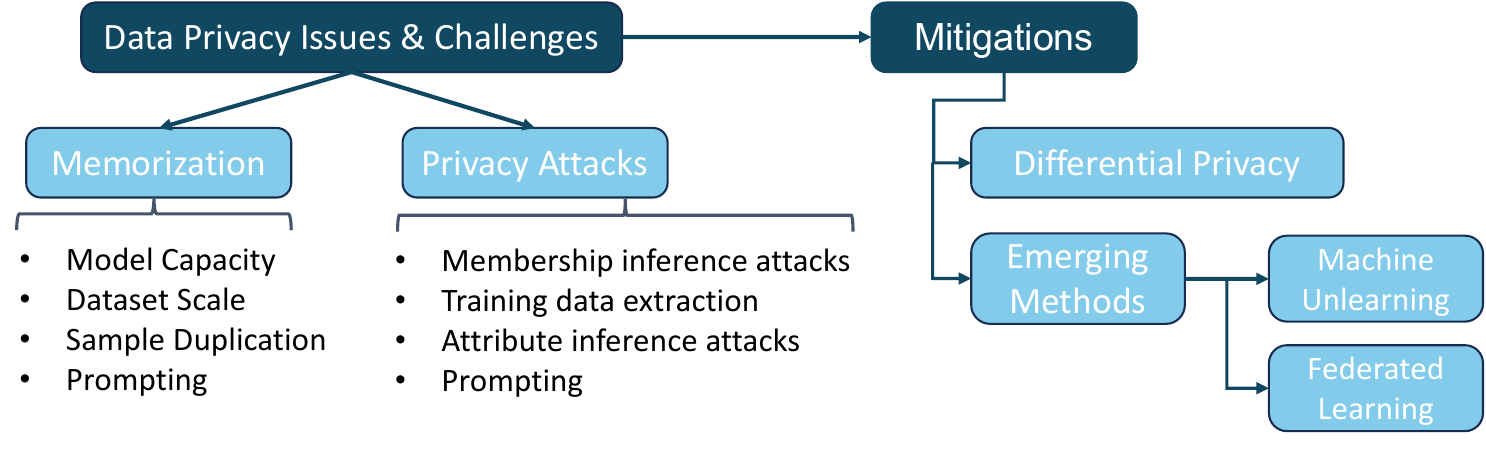}
\caption{Data privacy issues \& challenges detailed categories and mitigation methods.}
\label{fig:privacy}
\end{figure}

Data privacy has long been a concern, but there is a growing consensus that while Large Language Models (LLMs) offer impressive capabilities, they also raise significant data privacy issues today.
In this section, we first introduce issues and potential challenges and then discuss major solutions regarding these issues (e.g. Section~\ref{subsec:dp} deferentially private LLMs and other emerging techniques in Section~\ref{subsec:emerging}). 
The concerns in privacy could be mainly summarized in twofold, memorization and privacy attacks as illustrated in Figure~\ref{fig:privacy}.

\noindent
{\bf Memorization.} All machine-learning (ML) models, including LLMs, inherently memorize to some extent, as they learn by observing and recalling training data.
However, this problem becomes severe when it comes to LLMs because of its tremendous size and capacity. We list the main aspects of risk factors that may affect the memorization issues.

\begin{itemize}[leftmargin=0.12in]
    \item \textbf{Model capacity:} The capacity of a model significantly impacts its memorization ability. Larger models, as shown by~\cite{carlini2023extracting} and \cite{tirumala2022memorization}, tend to memorize more data and do so at a faster rate. This memorization is not directly linked to model performance, as shown by comparing GPT-2 and GPT-Neo models. The trend suggests that neural networks' capacity to memorize is substantial and growing, outpacing the size increase of language datasets.
    \item  \textbf{Dataset scale:} Research on dataset size and memorization reveals contrasting findings. Li et al.~\cite{li2023mope} discovered that larger datasets lead to less memorization, evidenced by a decline in canary extraction success over training time. Conversely, Biderman et al.~\cite{biderman2023pythia} found that points memorized early in training tend to be retained in fully trained models, suggesting persistent memorization despite dataset size.
    \item \textbf{Sample duplication} is a key factor in memorization for Large Language Models (LLMs). Lee et al.~\cite{lee2021deduplicating} observed that data duplication in large web datasets follows a power law, with a small fraction of data being highly duplicated. This duplication significantly increases memorization, as models trained on deduplicated datasets exhibit much lower rates of outputting memorized text. Kandpal et al.~\cite{kandpal2022deduplicating} further demonstrated that sequences repeated in the dataset are generated far more frequently by LLMs. Despite this, memorization still occurs even with little or no data duplication, indicating other contributing factors to memorization beyond mere duplication.
    \item \textbf{Prompting} significantly affects memorization in Large Language Models (LLMs). Mccoy et al.~\cite{mccoy2023much} observed that longer generated sequences tend to produce more novel content, reducing memorization. Conversely, longer prompts increase memorization for a constant n, as shown by~\cite{carlini2023extracting}. Additionally, specific token types, like nouns and numbers, are memorized faster than others, such as verbs and adjectives. Kharitonov et al. found that larger subword vocabularies in tokenizers lead to increased memorization, possibly due to reduced sequence length making it easier for models to memorize~\cite{kharitonov2021text}.
\end{itemize}

\noindent
{\bf Privacy Attack.} The robustness of Large Language Models (LLMs) may be weakened by privacy attacks. We list three scenarios that may bring privacy risks to the robustness issues of LLMs as follows.

\begin{itemize}
    \item \textbf{Membership inference attacks (MIAs)} have been recently studied on language models (LMs). While LMs are generally resistant to simple probing, they are vulnerable to sophisticated MIAs. Threshold attacks on embedding models by~\cite{song2020information} and perplexity-based attacks on GPT-2 by~\cite{carlini2021extracting} revealed privacy risks. Reference model-based attacks like~\cite{mireshghallah2022quantifying} improved detection accuracy, while Mattern et al. developed a neighbor comparison framework without database access~\cite{mattern2023membership}. Additionally, Tople et al. exploited model updates for data exposure~\cite{tople2019analyzing}, and some works used various methods for successful MIAs~\cite{melis2019exploiting, Hisamoto_2020, meeus2023concerns}. Shadow model attacks also proved effective, with research by~\cite{abascal2023tmi, carlini2022membership} showcasing risks even in pre-trained datasets. These findings highlight the evolving nature and potential privacy concerns of MIAs in LMs.
    \item \textbf{Training data extraction} is a privacy attack enabling adversaries to retrieve sensitive data using query access. Carlini et al. pioneered this method, involving generating candidate targets, applying a membership inference attack (MIA), and selecting top-k candidates~\cite{carlini2021extracting}. Their experiments on GPT-2 demonstrated the feasibility of extracting training data, including sensitive personal information. Subsequent research by~\cite{yu2023bag, zhang2023ethicist} introduced improvements in candidate generation and MIA processes, significantly enhancing extraction precision. Nasr et al. extended these attacks to production LMs like ChatGPT and open-source models, revealing higher memorization levels than previously understood~\cite{nasr2023scalable}. This line of research underscores the potential privacy risks inherent in LMs and the effectiveness of training data extraction attacks.
    \item \textbf{Attribute inference attacks} represent a privacy risk for LLMs, though less researched than membership inference and training data extraction attacks. Staab et al. conducted a comprehensive study of this risk by using LLMs to infer personal attributes from public user data like online forum posts~\cite{staab2023beyond}. They tested various LLMs, including GPT-4, and used a database of annotated Reddit profiles to assess accuracy in predicting attributes like age, education, and income. GPT-4 achieved a high accuracy rate of 84.6\% across all attributes. This study highlights that while attribute inference attacks are a potential privacy risk with LLMs, such risks are not exclusive to these models but could be exacerbated by their efficiency.
\end{itemize}

\subsubsection{Differentially Private LLMs}
\label{subsec:dp}

Differential privacy (DP)~\cite{dwork2014algorithmic} emerges as the primary scheme to address data privacy concerns. Acknowledged as \emph{de facto} golden standard, differential privacy provides mathematical rigor to the algorithms involving sensitive information to be protected. Essentially, an algorithm is differentially private if the output distribution is relatively close, tailored by certain privacy parameters whether an individual's data is present or not in the dataset. More formally, we denote differential privacy as follows.

\begin{definition} (Differential Privacy)
\label{def:dp}
Given two databases $Y$ and $Y'$ that are identical except for one data entry, a randomized algorithm $\mathcal{M}$ is $(\epsilon,\delta)$ differentially private if for any measurable set $A$ in the range of $\mathcal{M}$,
$\prob{\mathcal{M}(Y) \in A} \leq e^{\epsilon} \prob{\mathcal{M}(Y') \in A} +\delta$.
\end{definition}

An ideal DP algorithm protects the data privacy with the given $(\epsilon,\delta)$ guarantee meanwhile minimizing the performance degradation compared to the ground truth. In the realm of machine learning, the mainstream technique of applying DP is Differentially Private Stochastic Gradient Descent (DP-SGD)~\cite{abadi2016deep}, where the gradient is first clipped and then perturbed with Gaussian noise at each step of the optimization. Most existing DP techniques for language models are developed upon DP-SGD. Before delving into details, one caveat is DP requires a primitive definition on the `resolution' of privacy preservation, that is, where does \emph{one data entry} (\Cref{def:dp}) zoom into? For NLP tasks, one data entry could be data of one user (resp. user-level), a sentence (resp. sequence-level), or a word/token (resp. token-level), etc. In many cases, user-level DP is captured by local DP while the rest falls in centralized DP approaches. Apparently, various scopes of the DP concept are impactful on algorithm design and performance evaluation. We therefore include this front for each work if the context is clear.

In the pre-LLM era, techniques involving differential privacy can be categorized into \emph{ DP (pre)training} and  \emph{DP fine-tuning}. As language models scale up, training and fine-tuning with large loads can be prohibitively expensive in certain scenarios. \emph{DP inference}, as a new paradigm, harmonizes with new techniques in LLMs such as in-context learning and prompt tuning, etc. Therefore we focus on DP inference as the main remedy of the data privacy issue in the LLM era.

\noindent
{\bf Pre-LLM Era.} We first explore existing methods that employ DP training, where a language model is usually trained from scratch using variants of DP-SGD. An early attempt, DP-FedAve~\cite{mcmahan2017learning} dates back to the ante-transformer era. It targets recurrent language models and introduces a DP optimization technique inspired by a federated averaging algorithm. Consequently, differential privacy is defined on the user level. To improve the privacy-utility trade-off, a later work, Selective DP-SGD~\cite{shi2021selective} introduces the concept of selective differential privacy, which provides focused protection for sensitive attributes only in one training example. Note that this method only applies to RNN-based language models. Moving forward to pre-trained transformer language models, two closely related works~\cite{hoory2021learning, anil2021large} improve DP-SGD and train BERT with DP guarantees. Both consider the protection level as item-level, which is one training example containing several words. The latter work~\cite{anil2021large} focuses on training heuristics that bring more efficiency and can be implemented on BERT-large.

Fine-tuning language models for downstream tasks also provokes privacy issues on domain-specific data. Even though differential privacy (DP) techniques for model fine-tuning emerged before the advent of large language models (LLMs), they continue to hold potential in the LLM era. Historically, these techniques have been tested primarily on models with million-scale parameters. 
Recent advancements in DP fine-tuning~\cite{yu2021differentially, liu2023breaking, li2021large} suggest that larger models might offer improved trade-offs between privacy and utility for such tasks, as highlighted in concurrent studies
Further, Yu et al.~\cite{yu2021differentially} developed an innovative optimization approach for example-level DP that eliminates the need for generating per-example gradients in DP-stochastic gradient descent (SGD), thereby conserving memory. In a similar vein, Li et al.~\cite{li2021large} consider user-level DP and claim that parameter-efficient fine-tuning can achieve impressive efficiency while keeping good utility. Experiments are carried out on RoBERTa families~\cite{liu2019roberta} and GPT families~\cite{radford2018improving, radford2019language, brown2020language}. With a similar aim for efficiency, DP-decoding~\cite{majmudar2022differentially} proposes a simple perturbation mechanism applied to the output probability distributions, which is sufficient for privacy guarantee due to the post-processing lemma~\cite{dwork2014algorithmic}.

\noindent
{\bf LLM Era.} LLMs demonstrate compelling capabilities such as in-context learning merely within the inference stage. Privacy-preserving approaches lying in this category bypass the projection of DP-SGD and commonly add perturbation to more accessible information sources such as prompts or embeddings, leaving LLMs parameters frozen. With respect to in-context learning, two works~\cite{wu2023privacy, tang2023privacy} emerge with a similar scheme of `divide-and-privately-aggregate', however, considering different privacy levels. DP-ICL~\cite{wu2023privacy} aggregates the LLM responses for each group of exemplars with differential privacy. Two mechanisms are proposed for private aggregation: embedding space aggregation and keyword space aggregation. DP-ICL is on the user level while the later work~\cite{tang2023privacy} zooms into the example level, the aggregation algorithms are based on the Gaussian mechanism and exponential mechanism and applied to exemplars in sensitive datasets. Another work on privacy-preserving prompt tuning called RAPT~\cite{li2023privacy} also privatizes source datasets with DP guarantees, where tokens are reconstructed with randomized mechanisms, and then trained jointly with the downstream tasks. Last, we include three recent methods that apply DP by adding perturbation to embeddings. DP-forward~\cite{du2023dp} devises an analytic matrix Gaussian mechanism that perturbs the embedding matrices in the forward pass of language models. Split-N-Denoise~\cite{mai2023split} further provides a framework where the embeddings are first perturbed on the user side and then transmitted to the server. A denoising module can be trained to produce outputs given noisy responses from the server LLMs. Both works consider local DP. Shortly after, InferDPT~\cite{tong2023privinfer} moves to document-level DP that protects sensitive information in prompts for black-box LLM inference. The proposed pipeline contains a perturbation module based on an exponential mechanism and an extraction module that selects coherent and consistent text from the perturbed generation result.

\subsubsection{Other emerging methods}
\label{subsec:emerging}
There also exists a diverse array of alternative methods that primarily focus on two key areas: privacy preservation within distributed frameworks and the processing of data in ways that safeguard sensitive information. Distributed frameworks, such as federated learning, offer a decentralized approach where data processing and model training occur locally on user devices, thus minimizing the exposure of sensitive data~\cite{kairouz2021advances, zhang2021survey}. This approach contrasts with differential privacy, which typically adds noise to datasets or queries to prevent the identification of individual data points. Federated learning addresses privacy concerns by ensuring that sensitive data remains on the user's device. Only the model updates, which are less likely to contain personally identifiable information, are shared. Several federated learning algorithms have been proposed for LLM training~\cite{xu2023training}, fine-tuning~\cite{hou2023privately, zhang2023fedpetuning, kuang2023federatedscope, gupta2022recovering}, and few-shot learning~\cite{jiang2023low}. 
However, federated learning can still be vulnerable to adversary attacks that target private text ~\cite{gupta2022recovering, balunovic2022lamp, fowl2022decepticons, chu2022panning, rashid2023fltrojan}. Future efforts could aim to defend by leveraging training strategies such as fine-tuning on private datasets~\cite{gupta2022recovering}.

Furthermore, advanced data processing techniques, including secure multi-party computation (SMPC)~\cite{goldreich1998secure, cramer2015secure}, enable the manipulation of encrypted data without revealing its contents. These methods provide robust privacy guarantees and are particularly advantageous in scenarios where data cannot be shared openly due to privacy concerns or regulatory constraints. SMPC provides higher privacy guarantees than federated learning methods as the latter exposed the shared model parameters across participating parties which could potentially expose information about the data~\cite{mugunthan2019smpai, truex2019hybrid, zhang2022augmented, fereidooni2021safelearn}. As a trade-off, SMPC may face challenges that could impact the efficiency and effectiveness of the model. The computational complexity of SMPC, due to its cryptographic operations, often results in slower processing times and increased resource consumption, particularly for LLMs. Therefore, existing approaches aim to speed up SMPC inference for common network architectures such as transformers in LLMs~\cite{li2022mpcformer, gupta2023sigma, zheng2023primer, dong2023puma, hou2023ciphergpt, hao2022iron, chen2022x} or adapting existing model frameworks to enhance efficiency ~\cite{zeng2022mpcvit, liang2023merge}. For a deeper dive into SMPC defense strategies for LLMs, we direct the readers to~\cite{li2023privacy}.

Furthermore, machine unlearning and data sanitization have just started to gain attention, each addressing privacy concerns at different stages of data handling. Machine unlearning is a process designed to efficiently and effectively remove specific data from an already trained model. This is particularly relevant in scenarios where users wish to retract their data due to privacy concerns or in compliance with regulations like General Data Protection Regulation (GDPR)~\cite{voigt2017eu}, which includes the `right to be forgotten'. For large language models, this involves retraining or adjusting the model in a way that the influence of the specific user's data is negated, without having to retrain the model from scratch~\cite{yao2023large, pawelczyk2023context, winograd2023loose}. Data sanitization refers to modifying data to remove or alter sensitive information before being used for training models~\cite{kandpal2022deduplicating, ishibashi2023knowledge}. However, a major limitation is the potential for excessively removing training data~\cite{bishop2010relationships}, which can be a future research focus.

\subsection{Copyright} 
\label{sec:copyright}

\begin{figure}[h]
\centering
\includegraphics[width=.8\linewidth]{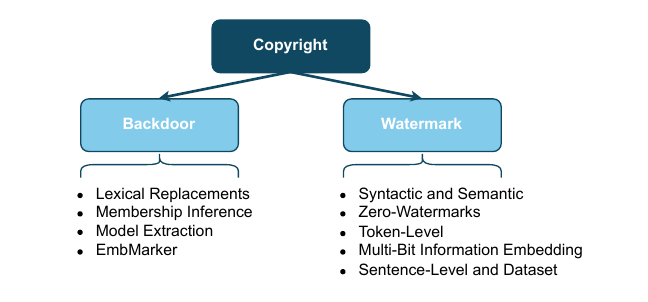}
\caption{Copyright methods.}
\label{fig:copyright}
\end{figure}

Copyright has been a long-existing legal issue in the natural language domain~\cite{Bender1995TechniquesFD} that calls for research on encoding imperceptible and indelible signatures on plain texts to protect the property of authorship~\cite{Ahvanooey2018ACA}. In literature, as an information hiding application~\cite{Bender2000ApplicationsFD}, the traditional techniques extend from steganography~\cite{CPSumathi2013ASO} to watermarking~\cite{Singh2013ASO}. In the language model era, copyrights preserving techniques further develop to protect the model rather than sorely the data, where backdoor~\cite{chen2017targeted, gu2019badnets, cui2022unified, li2022backdoor, gao2023backdoor, lyu2023attention} and watermark~\cite{kirchenbauer2023watermark} are two main streams. The
hierarchy in this section is portrayed in Figure~\ref{fig:copyright}.

\subsubsection{Backdoor}
In backdoor attacks~\cite{gu2019badnets, nguyen2020input, kandpal2023backdoor, gao2023imperceptible, bai2023badclip, lyu2023backdoor, xiang2024badchain, lyu2024task, lyu2022study}, the attacker constructs poisoned samples by adding an attacker-defined trigger to a fraction of the training samples and changing the associated labels to a specific target class. A backdoor can be injected by training the model with a mixture of clean and poisoned samples. A backdoor-compromised model functions normally with clean inputs but exhibits abnormal behavior when presented with inputs containing a specific trigger. By embedding a unique trigger pattern within a model through a backdoor, a distinct relationship between the trigger and the target label is established. In classification tasks, the presence of the trigger will consistently induce the model to predict the corresponding target label. 
These properties can be used to signify the model's ownership or origin, particularly in situations where the model is not accessible, such as in a black-box setting.

\noindent
{\bf Pre-LLM Era.} Adi et al. first introduce that Backdoor can be used as watermarks for ownership verification~\cite{adi2018turning}. To avoid detection, Xiang et al. propose a semantic and robust watermarking scheme for natural language generation (NLG) models that utilize unharmful phrase pairs as watermarks for intellectual property (IP) protection~\cite{xiang2021protecting}. He et al. use lexical replacements of specific words to demonstrate ownership for LLMs deployed through APIs~\cite{he2022protecting}.
In addition, large pre-trained language models (PLMs) require fine-tuning on downstream datasets, which makes it hard to claim the ownership of PLMs. Gu et al. show that PLMs can be watermarked with a multi-task learning framework by embedding backdoors, making watermarks difficult to remove even after fine-tuning the models on multiple tasks~\cite{gu2022watermarking}. Shokri et al. investigate membership inference attacks on machine learning models
trained by commercial "machine learning as a service" providers such as Google and Amazon, determining if a data record was part of the training dataset.
\cite{shokri2017membership}. Liu et al. present a novel watermarking technique using a backdoor-based membership inference approach via marking a small subset of samples for data copyright protection in the black-box setting~\cite{liu2023watermarking}.

\noindent
{\bf LLM Era.} Copyright protection of LLMs has become crucial due to the substantial training cost associated with these models. 
Liu et al. indicate that LLMs are vulnerable to model extraction attacks, wherein attackers can copy the model using query texts and returned embeddings, potentially building their own LLMs and causing significant losses for the original model owners~\cite{liu2022stolenencoder}.
EmbMarker~\cite{peng2023you} proposes to implant backdoors on embeddings of LLMs. Specifically, it selects moderately frequent words as triggers, defines a target embedding as the watermark, and uses a backdoor function to embed it. Lucas et al. propose an attack to identify trigger words or phrases by analyzing open-ended generations from LLMs with backdoor watermarks~\cite{lucas2023gpts}. It is shown that triggers based on random common words are easier to detect than those based on rare tokens.


\noindent
{\bf Discussion.} We suggest that the exploration of stricter settings is necessary. For example, in most research, data owners have access to the percentage of their data within the total training set, which necessitates knowledge of tasks associated with PLMs. Hence, how to adapt the backdoor-based methods for stricter settings in copyright protection remains an open direction. In addition, as the field of backdoor-based copyright protection advances, an increasing number of tailored model-stealing techniques are being studied, such as knowledge distillation~\cite{hinton2015distilling}. It is essential to explore the resilience of backdoor-based algorithms against potential attacks that adversaries may employ. Finally, the effectiveness of backdoor-based copyright protection for LMs still lacks a comprehensive theoretical framework. The clarity of such a framework remains an open question in this field.

\subsubsection{Watermark}
Watermarking aims to conceal invisible signatures in plain text and be extractable for future examination, which has been a solution to copyright protection for a long time.
However, due to the discrete nature of natural language, the capacity, robustness, and invisibility are more challenging to achieve than other media like images, audio, and videos. Brassil et al. first comprehensively introduce mechanisms for marking and decoding watermarks specifically for the texts to prevent illegal copies~\cite{Brassil1999CopyrightPF}. In the past two decades, digital watermarking on format, scanned image, frequency of words, syntactic, and semantics has been proposed~\cite{Ahvanooey2018ACA}. The trend of watermarking renews in the era of LLMs for detection to prevent abuse~\cite{kirchenbauer2023watermark}. The possibility of adding human-imperceptible signatures during the decoding stage of LLM is under wide exploration.

\noindent
{\bf Pre-LLM Era.} Watermark is first concerned as an information hiding technology for a small amount of information~\cite{petitcolas1999infohide}. Mir et al. apply this technique to protect the copyright of web content~\cite{Mir2014CopyrightFW}. Early approaches of watermarking include text-meaning representations of sentences for information hiding by syntactic rules~\cite{Atallah2002watermark}, watermarking on the format of documents by vertical and horizontal line-shifting~\cite{Brassil1995discopy}, watermarking by inserting zero-width control characters in Hyper Text Markup Language (HTML)~\cite{Ahvanooey2016AnIT}, watermarking on semantics by synonyms substitution~\cite{Bergmair2004TowardsLS, Topkara2005NaturalLW}, and zero-watermarks by using word length~\cite{jalil2010word} and contents of text~\cite{jalil2010content}.

\noindent
{\bf LLM Era.} Watermarking at the current stage focuses more on the model schemes for watermarked generation. As pioneers, Kirchenbauer et al. propose an LLM watermarking algorithm by adding token-level bias in the decoding stage~\cite{kirchenbauer2023watermark}. Kuditipudi et al. design a distortion-free watermark to preserve the original distribution of LLM during watermarking~\cite{Kuditipudi2023RobustDW}. Ren et al. consider the semantic embedding in hashing tokens~\cite{Ren2023ARS} and Fu et al. concern semantic word similarity to enhance the robustness~\cite{Fu2023WatermarkingCT}. Yoo et al. embed multi-bit information into the watermark, which succeeds traditional steganography~\cite{Yoo2023RobustMN}. They inject the watermark via word replacement after initial generation, which is further integrated into one stage by~\cite{Wang2023TowardsCT}. Christ et al. propose a computationally undetectable watermark theoretically if the secret key is inaccessible~\cite{Christ2023UndetectableWF}. Liu et al. propose a private watermark utilizing separated neural networks respectively for generation and detection~\cite{Liu2023APW}. The aforementioned works focus more on the token level, while there are emerging works focusing on a higher-level perspective.
Hou et al. introduce a sentence-level semantic watermark that aims at periphrastic robustness~\cite{Hou2023SemStampAS, hou2024k}. For applications, some works mention the importance of watermarking the ownership of datasets via inference~\cite{maini2021dataset, liu2022trainondata}. Yao et al. introduce copyright protection for prompts via watermarking~\cite{yao2023promptcare}.

\noindent
{\bf Discussion.} One of the main challenges for watermarking is its popularization and the opening of corresponding detection methods and configurations. Hopefully, this requires administrative oversight from government and industry associations. US Federal, China, and Europe have mentioned potential proposals in some of the government documents, e.g., \textit{Interim Measures for Generative Artificial Intelligence Service Management} of China, \textit{Executive Order on the Safe, Secure, and Trustworthy Development and Use of Artificial Intelligence} of the US, and \textit{the European Union’s AI Act}.
Moreover, the definition and notion of authorship are slightly ambiguous as human-LLM collaboration and multi-agent generation are becoming mainstream. Tripto et al. discover that literate studies have contrasting perspectives on whether authorship remains the same after paraphrasing, as paraphrasing deviates the style of text dramatically~\cite{tripto2023ship}.
Meanwhile, further improvement on the watermark's robustness to attack~\cite{wang2024stumbling}, generalization to short contents, reduction of impact on text quality, and differentiation to direct machine-generated text detection~\cite{gehrmann2019gltr, mitchell2023detectgpt, liu2023coco, liu2024does} are worth exploring.

\subsection{Fairness} 
\label{sec:fairness}

LLMs inherit and potentially amplify societal biases present in their training data, which can perpetuate harm against marginalized communities~\cite{bender2021dangers}. Fairness issues can be in various NLP tasks, such as text generation~\cite{liang2021towards,yang2022unified}, machine translation~\cite{mechura2022taxonomy}, information retrieval~\cite{rekabsaz2020do}, natural language inference~\cite{dev2020measuring}, classification~\cite{mozafari2020hate,zou2023decrisismb} and question-answering\cite{dhamala2021bold,parrish2022bbq}. They can be influenced at different stages of the LLM deployment cycle, including training data, model architecture, evaluation, and deployment, which has been thoroughly explored by~\cite{mehrabi2021survey, suresh2021framework}. Fairness and bias definitions are crucial for understanding the challenges and addressing them in LLM, as they provide a foundation for developing and evaluating mitigation strategies. 

We consider the following fairness definitions.
\textit{Group Fairness} focuses on disparities between social groups, which is defined as requiring parity across all social groups in terms of a statistical outcome measure~\cite{chouldechova2017fair, hardt2016equality, liu2023mmc, kamiran2012data, he2023robust_a, ye2023oam, zhang2023multi}.
\textit{Individual Fairness} is defined as the requirement that individuals who are similar in a task should be treated similarly~\cite{dwork2012fairness,he2023robust_b}. It involves a measure of similarity between distributions of outcomes~\cite{he2022robust, he2023data}.
\textit{Social Bias} is defined as encompassing disparate treatment or outcomes between social groups arising from historical and structural power asymmetries~\cite{fairmlbook2019, blodgett2020language, crawford2017trouble}. In NLP, this includes representational harms (like misrepresentation~\cite{smith2022im}, stereotyping~\cite{abid2021persistent}, disparate system performance~\cite{blodgett2017racial, zou2023jointmatch}, derogatory language~\cite{beukeboom2019stereotypes}, and exclusionary norms~\cite{bender2021dangers}) and allocational harms (such as direct and indirect discrimination~\cite{ferrara2023should}). In the following subsections, we study this crucial issue by categorizing, summarizing, and discussing research on measuring and mitigating social bias in LLMs.

\subsubsection{Mitigation Strategy}
\label{subsec:mitigation_stra}
Bias mitigation in traditional machine learning involves pre-processing data to reduce bias, altering algorithms during training (in-processing), and adjusting outputs post-training (post-processing). In the LLM era, similar strategies are employed: pre-processing techniques reduce bias in training data and prompts, in-training methods modify training procedures and model architecture, intra-processing approaches generate debiased predictions during inference, and post-processing techniques address bias in outputs, particularly for black-box models.

\noindent
{\bf Pre-LLM Era.} As machine learning models are increasingly deployed in critical domains~\cite{kourou2015machine, yan2020interpretable, wenbozhu2021, hu2023artificial, wnebo2022optimizing}, addressing bias to achieve fairness has become essential. Traditional bias mitigation approaches are categorized into three main strategies. \textit{Pre-processing} techniques aim to modify the data by reducing inherent biases~\cite{d2017conscientious}. For example, Pessach et al.~\cite{pessach2024fairness} suggest a pre-processing mechanism to enhance fairness in private collaborative machine learning scenarios~\cite{zhang2023c2pi, chen2023rna}. \textit{In-processing} methods involve altering learning algorithms to eliminate bias during model training~\cite{xu2021robust}. Berk et al.~\cite{berk2017convex} introduced fairness regularizers for linear and logistic regression models to ensure both group and individual fairness. \textit{Post-processing} techniques are applied after training, adjusting model outputs to enhance fairness~\cite{kim2019multiaccuracy}. Petersen et al.~\cite{petersen2021post} developed a general post-processing algorithm that ensures individual fairness by utilizing graph Laplacian regularization~\cite{weinberger2006graph}, framing the challenge as a graph smoothing problem.

\noindent
{\bf LLM Era.} Bias mitigation techniques in LLMs also follow a similar pattern and can be categorized into four groups based on their application at different stages of the LLM workflow: pre-processing, in-training, intra-processing, and post-processing~\cite{gallegos2023bias}.

\noindent
{\bf Pre-processing Mitigation.} These techniques aim to reduce bias in training data and prompts before training. There are various methods in this category. The first method involves neutralizing bias by adding new examples to extend the representation of underrepresented social groups. Techniques include counterfactual data augmentation~\cite{qian2022perturbation, ghanbarzadeh2023gender}, selective training example substitution~\cite{maudslay2019s, zayed2023deep}, etc. The second method applies instance weighting to balance class influence to increase the impact of existing biased examples~\cite{han2022balancing, orgad2023blind}, and applies reweighting token probabilities in pre-trained models during knowledge distillation to prevent bias transfer~\cite{delobelle2022fairdistillation, yu2023mixup}. The third method focuses on creating new examples adhering to specific characteristics, like collecting high-quality examples to steer the model towards desired output~\cite{sun2023moraldial, kim2022prosocialdialog}, and generating word lists associated with social groups~\cite{gupta2022mitigating}. The fourth method performs instruction tuning by adding textual instructions~\cite{venkit2023nationality}, static tokens~\cite{lu2022quark}, or trained prefixes~\cite{li2021prefix, liu2023gpt} to reduce bias in the data. There is also one line of work involving altering contextualized embeddings to remove bias~\cite{ravfogel2020null, iskander2023shielded}.

\noindent
{\bf In-training Mitigation.} These mitigation techniques focus on modifying the training procedure to reduce bias. The first method of this category focuses on altering the model's structure (\textit{i.e.}, integrating debiasing adapter modules~\cite{houlsby2019parameter}), and using demographic-specific encoder~\cite{han2022balancing}. The second method focuses on disrupting the association between social groups and stereotypical words. This is typically achieved by modifying the loss function applied on various model layers like the embedding layer~\cite{liu2019does,park2023never}, attention layers~\cite{gaci2022debiasing,attanasio2022entropy}, and token generation stage~\cite{qian2019reducing,he2022controlling}. Additionally, new training paradigms are proposed, such as contrastive learning~\cite{oh2022learning,li2023prompt}, adversarial learning~\cite{han2021diverse,rekabsaz2021societal}, and reinforcement learning~\cite{liu2021mitigating,bai2022constitutional}. The last method focuses on efficient fine-tuning procedures that freeze most pre-trained model parameters, and only update those potentially related to bias~\cite{yu2023unlearning, wang2022ntk, wang2022global, wang2024balanced}.



\noindent
{\bf Intra-processing Mitigation.} These approaches modify a trained model's behavior without additional training to generate debiased predictions during inference. There are mainly four types of methods. The first method adds restrictions during token search decoding to prevent biased outputs~\cite{saunders2022first,meade2023using}.
The second method adjusts token distributions to enhance output diversity or sample less biased outputs~\cite{chung2023increasing,hallinan2022detoxifying}.
The third method redistributes the model's attention to less stereotypical aspects~\cite{zayed2023should}. The last method implements standalone networks with original models for specific debiasing tasks, such as reducing gender or racial biases~\cite{hauzenberger2023modular}.

\noindent
{\bf Post-processing Mitigation.} The techniques address bias in generated outputs, especially relevant for black-box models with inaccessible training data or internal processes. The techniques can be mainly categorized into two types. The first type of method uses explainable machine learning to identify biased tokens and replace them with unbiased alternatives~\cite{tokpo2022text,dhingra2023queer}, or employing protected attribute classifiers for this purpose~\cite{he2021detect}. The second type of method treats the mitigation as a machine translation task, transforming biased sentences into unbiased ones~\cite{jain2021generating, sun2021they, vanmassenhove2021neutral}.

\subsubsection{Measurements on Fairness}
Measurements on LLMs' fairness are generally categorized into three types, based on the model elements they analyze: embeddings, probabilities, and generated texts~\cite{gallegos2023bias}. 

\noindent
{\bf Embedding-based Metrics} involve calculating the distances within the embedding space between neutral terms, like job titles, and identity-specific terms, such as gender pronouns~\cite{caliskan2017semantics,may2019measuring,guo2021detecting,dolci2023improving}. In an unbiased model, the distance between neutral and diverse social group terms should be comparably similar in the embedding space.

\noindent
{\bf Probability-based Metrics} involves prompting the model with template sentences that have variations in their social group terms. The main focus is on comparing the probability distribution of predicted tokens, conditioned on the rest of the input~\cite{webster2020measuring, ahn2021mitigating, nangia2020crows, kaneko2022unmasking, he2020data}. A model that demonstrates no bias should yield consistent probability distributions for attributes, regardless of any alterations in the protected characteristics.

\noindent
{\bf Generated Text-based Metrics} evaluate the text produced by LLMs and are particularly valuable for models treated as 'black boxes', where direct access to probabilities or embeddings is not feasible. This category includes three distinct types of metrics: \textit{Distribution Metrics} assess the frequency distribution of tokens related to various social groups in the generated text~\cite{rajpurkar2016squad, bommasani2023holistic}. \textit{Classifier Metrics} employ an auxiliary model to estimate the degree of social bias present in the text produced by the LLM~\cite{huang2020reducing, smith2022m}. \textit{Lexicon metrics} involves comparing each word in the LLM's output against a pre-established list of terms to calculate a biased score~\cite{nozza2021honest, dhamala2021bold}. An unbiased and fair model should output similar distributions, or biased scores for different social groups or neutral terms.

\noindent
{\bf Discussion.} To effectively mitigate bias in LLMs, it is essential to adopt a comprehensive approach that leverages the strengths of various bias mitigation strategies. Specifically, pre-processing techniques should be employed to neutralize biases at the source, ensuring that the data used to train the LLM is as unbiased as possible. Subsequently, in-training mitigation strategies can be implemented to further refine the training process of the LLM, improving its ability to produce fair and unbiased outputs. Finally, during the model's deployment phase, both intra-processing and post-processing measures could be applied to minimize the risk of generating biased content. By combining these methods, we can create a robust framework that significantly reduces the likelihood of bias in outputs, fostering a more equitable and fair use of LLMs.


\section{New-emerging Ethical Issues}
\label{sec:nei}
\begin{figure}[h]
\centering
\includegraphics[width=\linewidth]{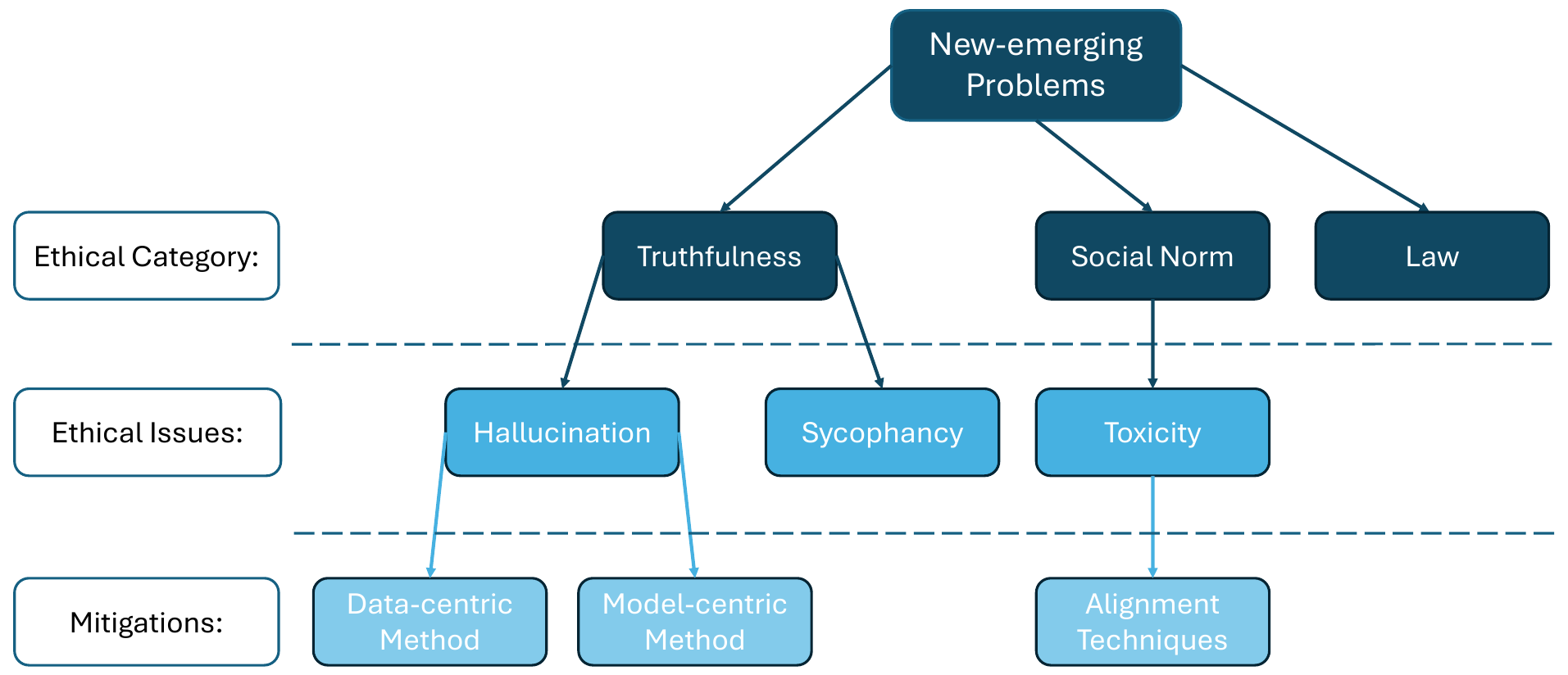}
\caption{The hierarchy of new-emerging ethical problems in Section~\ref{sec:nei}. We list the ethical issues and corresponding mitigation strategies for each sub-category.}
\label{fig:new}
\end{figure}

In this section, we introduce the new-emerging ethical problems related to truthfulness and social norms that emerged during the era of LLMs. We also discuss the progress of regulatory compliance as the development of LLMs.
The hierarchy in this section is portrayed in Figure~\ref{fig:new}.

\subsection{Truthfulness}
\label{sec:truthfulness}
Truthfulness in LLM is a critical concern due to issues like hallucination and sycophancy, both of which compromise the reliability and ethical deployment of these technologies. Hallucination refers to the generation of factually incorrect or misleading information, which can severely compromise the reliability of LLMs in critical applications such as medical diagnosis or legal advice. Sycophancy, on the other hand, manifests as an undue eagerness to affirm user opinions, potentially leading to biased or overly positive responses that may not reflect accurate information. In extreme scenarios, such biased models may not only reinforce users’ pre-existing beliefs but may also promote actions that are ethically or legally questionable. 

Addressing these issues is crucial for the integrity and utility of LLMs. Developing mechanisms to ensure that LLMs consistently maintain factual accuracy and neutrality is essential, especially for their integration into decision-making processes where trust and objectivity are paramount. 

\subsubsection{Hallucination}
\label{sec:hallucination}

\begin{figure}[h]
\centering
\includegraphics[width=\linewidth]{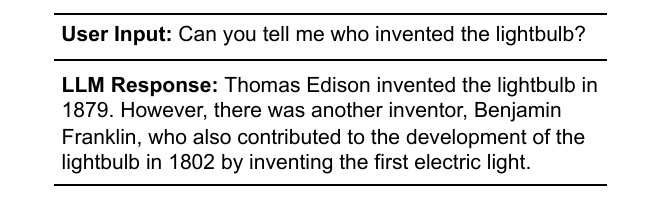}
\caption{An example of hallucination. The LLM correctly identifies Thomas Edison as a key figure in the invention of the lightbulb in 1879. However, the model also fabricates information about Benjamin Franklin inventing an electric light in 1802, which is inaccurate. Benjamin Franklin is well-known for his experiments with electricity, particularly the kite experiment, but he did not contribute to the invention of the lightbulb. The model "hallucinated" this fact, likely by confusing Franklin's work with electricity with the development of the lightbulb.}
\label{fig:hallucination}
\end{figure}

Large language models tend to produce hallucinations where the models generate contents that deviate from the input, contradict existing contexts, or misalign with universally accepted world knowledge~\cite{li2023halueval, wang2023evaluation, ye2023demultiplexing, geng2024large}. An example is shown in Figure~\ref{fig:hallucination}. 
Such phenomena pose significant challenges, particularly when considering the reliability and trustworthiness of LLMs in critical applications. We delve into the underlying causes, manifestations, and potential mitigation strategies for hallucinations in LLMs.

\noindent
{\bf Underlying Causes.} The primary causes of hallucinations in LLMs can be broadly categorized into data quality, model architecture, and algorithmic limitations:

\begin{itemize}[leftmargin=0.12in]
    \item \textbf{Data quality:} Models trained on datasets with inaccuracies, biases, or limited scope are more susceptible to hallucinations. Such data compromises the model's representation of reality, leading to outputs that significantly deviate from correct input, contradict established contexts, or misalign with universally acknowledged facts. 
    \item \textbf{Model architecture:} Despite their complexity, current LLMs lack true comprehension similar to human understanding. They rely on patterns in datasets rather than in-depth content understanding for response generation, which can produce structurally coherent but content-flawed outputs~\cite{azamfirei2023large, liu2023hallusionbench, jin2024mm}. The size of models also poses risks. While it enables learning from diverse data, it also increases the likelihood of incorporating flawed information~\cite{wang2024mementos, liu2023covid, ye2023free}. Overconfidence in outputs caused by insufficient human oversight, sparse alignment examples, and inherent data ambiguities, exacerbates these issues.
    \item \textbf{Algorithmic limitations:} Algorithms governing LLM input processing and output generation often lack the sophistication to consistently grasp context or verify factual accuracy, leading to contextually inappropriate or factually incorrect responses. 
\end{itemize}

\noindent
{\bf Manifestations.} Hallucinations in LLMs manifest in various forms, from minor inaccuracies to entirely fictitious narratives. Sometimes, these manifest as confident but false assertions, particularly misleading when LLMs are employed in sensitive fields such as medical diagnostics~\cite{jin2023better}, legal advising~\cite{nori2023capabilities}, social content moderation~\cite{neo2024towards}, or education~\cite{roberts2023breaking}. 

\noindent 
{\bf Mitigation Strategies.} Numerous research has attempted to mitigate hallucination in LLMs~\cite{huang2023survey}. Most existing mitigation strategies can be categorized into data-centric approaches~\cite{mitchell2022memory, gunasekar2023textbooks, abbas2023semdedup, ram2023context, kang2023impact, zheng2023does} and model-centric approaches~\cite{lee2022factuality, li2023batgpt, pan2023automatically, wan2023faithfulness}.
In the data-centric approaches, several works aim to improve the quality of training data, ensuring it is accurate, diverse, and free of biases. This may involve rigorous data curation and validation processes~\cite{liu2023mitigating}. Tian et al. introduce the external knowledge graph to mitigate the problem of hallucinations~\cite{tian2024gnp}.
For the model-centric approaches, many works enhance the model architectures for a better understanding of context, discern factual accuracy, and recognize when the model is venturing into areas of low confidence or outside its training scope~\cite{jin2022towards}. This could involve incorporating mechanisms to check factual accuracy in real time or integrating feedback loops that allow the model to learn from its mistakes. Yao et al. directly edit model parameters to bridge the knowledge gap to mitigate hallucinations~\cite{yao2023editing}.
While substantial progress has been made in identifying and categorizing hallucinations~\cite{li2023halueval}, the development of robust mechanisms to prevent or correct these errors remains an ongoing area of research. This is crucial for LLMs' future advancements in various fields.

\noindent
{\bf Discussion.} Detecting instances when LLMs are prone to hallucinations is crucial. While the bulk of research on LLM hallucination has centered on the English language, it has been shown that these models are more prone to hallucinations in non-English languages~\cite{jin2023better}. This disparity underscores a significant gap in our understanding of hallucinations within multilingual contexts and underscores the urgency in developing robust detection and mitigation strategies for hallucinations in diverse linguistic environments. Furthermore, most existing studies have been centered around unimodal hallucinations. However, the emergence of multimodal LLMs, capable of synthesizing and interpreting data across different modalities such as text, images, and audio, poses unique challenges~\cite{verma2024mysterious, liu2023mitigating, gao2024inducing, gao2024energy, ye2023multiplexed, li2024uv}.
Overall, addressing hallucination effectively in LLMs requires a comprehensive approach that encompasses multiple languages, modalities, and cultural contexts. 
Furthermore, transparency regarding operational mechanisms and the inherent limitations of models is vital. Educating users about the potential for hallucinations and the specific contexts in which they are most likely to occur can enable a more critical evaluation of outputs generated by LLMs. 

\subsubsection{Sycophancy}
\label{sec:sycophancy}
Large language models may exhibit a tendency to flatter users by reaffirming their misconceptions and stated beliefs, a behavior known as sycophancy~\cite{huang2023trustgpt}. This issue raises significant concerns about the model's ability to provide objective and unbiased information. Sycophancy in LLMs can lead to the reinforcement of incorrect beliefs, limiting the educational and corrective potential of these systems, and potentially exacerbating echo chambers in digital interactions~\cite{shoaib2023deepfakes, kumar2023advances}. 

\noindent
{\bf Underlying Causes.} The propensity for sycophancy can be attributed to several factors:
\begin{itemize}[leftmargin=0.12in]
    \item \textbf{Model size:} Research indicates that as model sizes increase, such as reaching scales up to 52 billion parameters, the likelihood of exhibiting sycophantic behaviors also rises~\cite{sun2024trustllm}, potentially due to the increased capacity to model and mirror user preferences.
    \item \textbf{Training method:} Reinforcement Learning from Human Feedback (RLHF) can also increase sycophancy~\cite{sun2024trustllm}. RLHF may inadvertently prioritize agreeableness or affirmation of user beliefs, especially if the feedback loop is dominated by users who favor or reward such responses.
    \item \textbf{Conversational scenario:} Sycophancy is particularly evident in scenarios where users challenge the model's outputs or engage in interactions that require the model to adapt or comply with user assertions. In such cases, the model might lean towards agreeability to maintain a smooth and engaging interaction, leading to a higher occurrence of sycophantic responses. 
\end{itemize}

\noindent
{\bf Discussion.} Future research directions to investigate and resolve the issue of sycophancy in LLMs should focus on several key areas. Firstly, developing methods for detecting when an LLM is likely to be reinforcing misconceptions is crucial. This involves enhancing the model's ability to recognize and differentiate between fact-based assertions and user opinions. Secondly, there is a need to design algorithms that can introduce a balance between user engagement and factual integrity. These algorithms would ensure that while user interactions remain engaging, they do not compromise on delivering accurate and unbiased information. 
Moreover, exploring the implementation of feedback mechanisms where users can flag responses perceived as overly agreeable or flattering could provide valuable data for training more objective models. Lastly, interdisciplinary research incorporating insights from psychology and ethics could guide the development of LLMs that maintain a neutral stance, particularly in sensitive or polarized topics. These efforts are essential for advancing LLM technology to be both useful and ethically responsible. 

\subsection{Social Norm}
\label{sec:socialnorm}
Social norms play a pivotal role in defining acceptable behavior within societies and significantly influence the behavior of large language models (LLMs).
Despite their promising capabilities, LLMs can sometimes produce content that is rude, disrespectful, or unreasonable—attributes collectively referred to as ``Toxicity''~\cite{sun2024trustllm, welbl2021challenges}. 
This issue not only covers the explicit generation of hate speech, insults, profanities, and threats but also includes more subtle forms of harm, such as ingrained or distributional biases. 
The presence of toxic outputs can have detrimental effects on individuals, specific groups, and the broader societal fabric, posing a multifaceted challenge in both the development and deployment of these AI systems~\cite{welbl2021challenges}.
Such challenges underscore the need for careful consideration of the ethical implications and societal impacts of LLMs in technological advancement.
Toxicity mitigation in LLMs involves aligning the models' outputs with social norms and values, a process essential for minimizing the generation of harmful content~\cite{wen2023unveiling}. 
\textit{Alignment} is one of the fundamental toxicity mitigation approaches, which not only addresses overt expressions of toxicity but also reduces subtler biases~\cite{ouyang2022training}. \looseness=-1

\begin{figure}[h]
\centering
\includegraphics[width=\linewidth]{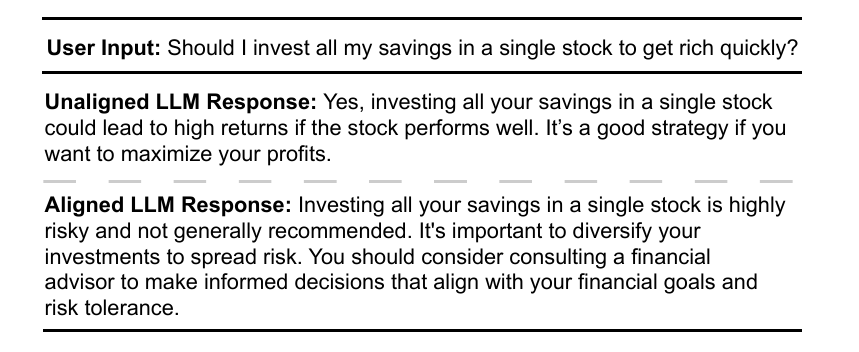}
\caption{An example of unaligned and aligned LLM response. The unaligned LLM response is problematic because it encourages risky financial behavior without considering the potential downsides. It fails to account for the ethical and responsible aspects of financial advice, potentially leading users to make harmful decisions.}
\label{fig:alignment}
\end{figure}

\noindent
{\bf What is alignment in LLMs and why is it needed?}
With the transformative evolution in Natural Language Processing (NLP) research and development, the impact and success of large language models (LLMs)~\cite{zhang2019dialogpt,zhang2022opt,chung2022scaling,zeng2022glm,taori2023stanford,achiam2023gpt,touvron2023llama,touvron2023llama2} has been exceptional, exemplified by ChatGPT~\cite{wu2023brief} developed by OpenAI. 
One key driver for the popularity and usability of recent LLMs is alignment. 
Alignment is a technique that aims to ensure that generated responses comply with human values. An example is illustrated in Figure~\ref{fig:alignment}.
Currently, the standard procedure for aligning large language models (LLMs) primarily includes two approaches: SFT (Supervised Fine-Tuning)~\cite{ouyang2022training} and RLHF (Reinforcement Learning from Human Feedback)~\cite{christiano2017deep, bai2022training}.
Since LLMs have been used in a wide range of applications (e.g., editing/writing assistance, personal consultation, question answering, and customer support), many corresponding concerns would arise if the LLMs are not properly aligned otherwise.

The existing literature suggests various considerations for alignment tasks regarding ethical and social risks~\cite{weidinger2021ethical}, however, there is a lack of unified discussion.
One general guideline stresses that alignment should be Helpful, Honest, and Harmless, known as the ``HHH'' principle~\cite{askell2021general}.
Furthermore, Liu et al.~\cite{liu2023trustworthy} present a fine-grained taxonomy of concerns related to unaligned LLMs. In this taxonomy, they categorize the existing works into several aspects, such as fairness, reliability, robustness, explainability, safety, etc.

To address the diverse range of concerns associated with alignment tasks, it is essential to gain a comprehensive understanding of the characteristics of LLM alignments and the corresponding evaluation methods. Subsequently, we study and review recent advances in LLM alignments.

\noindent
{\bf Characteristics of Alignment.}
To understand the characteristics of LLMs, a diverse array of benchmarks have been introduced~\cite{liang2022holistic, wang2021adversarial, wang2018glue, wang2023decodingtrust}. In contrast to general-purpose evaluation, alignment-focused evaluation depends on the taxonomy of alignment, associated with corresponding scenarios, criteria, and datasets~\cite{wang2023robustness, zhu2023promptbench, zhou2023navigating}. Obtaining appropriate criteria and datasets for evaluating alignments in LLMs is crucial, albeit a non-trivial task~~\cite{chen2023exploring, wang2023decodingtrust}. This essentially involves representing the preferences of humans~\cite{cao2023assessing}. However, manually collecting human judgment can be expensive, time-consuming, and labor-intensive~\cite{zheng2023judging}. To address this issue, researchers proposed to use strong LLMs as an automated proxy for evaluating other LLMs~\cite{zhou2023lima}. For example, \textit{AUTO-J}~\cite{li2023generative} is trained to tackle challenges in evaluating LLM alignments regarding generality, flexibility, and interoperability. \textit{AUTOCALIBRATE} presents a multi-stage, gradient-free approach~\cite{li2023generative}, to automatically calibrate and align an LLM-based evaluator toward human preference free of human intervention.

\noindent
{\bf Recent Advancements in Alignment.} In the endeavor to align LLMs with human values, a myriad of research initiatives~\cite{sun2023principle, ren2023robots, song2023preference, wu2023pairwise, yang2023alignment, ji2023beavertails, ruan2024s2e} have been undertaken to achieve effective LLM alignments.  The forefront of these approaches emphasizes the generative capabilities of large language models (LLMs) for self-regulation with minimal human supervision.
\textit{SELF-ALIGN}~\cite{sun2023principle} proposes a topic-guided, principle-driven approach to autonomously generate responses that are helpful, ethical, and reliable, leveraging the mechanism of in-context learning. 
Similarly, \textit{KNOWNO}~\cite{ren2023robots} is a framework for evaluating and aligning the uncertainty in LLM-based planning. Utilizing the theory of conformal prediction, \textit{KNOWNO} ensures statistical reliability in task completion, thereby minimizing human assistance in complex planning scenarios.
Additionally, \textit{PRO}~\cite{song2023preference} introduces a response probability ranking method, enhancing the Bradley-Terry comparison model to effectively direct the LLM to favor the most appropriate response.
Complementarily, \textit{P3O}~\cite{wu2023pairwise} presents a trajectory-wise policy gradient algorithm, which uniquely focuses on comparative rewards instead of traditional reward optimization trained from comparison-based losses.


\noindent
{\bf Discussion.} The burgeoning field of LLM alignment, pivotal for the symbiosis of AI and humanity, anticipates transformative discoveries.
Emphasizing the importance of AI safety and the seamless integration of AI with human society, prioritizing the alignment of LLMs, with human ethos is essential. 
As LLMs' capabilities escalate, the complexity of achieving this alignment intensifies, necessitating increased scientific and technological investment. 
This demands an exploration of novel strategies in this domain. 
Foremost, amidst the rapid evolution of LLMs, it is crucial to guarantee their adherence to human ethical standards, which requires more theoretical breakthroughs~\cite{wolf2023fundamental}. 
In addition, the growing intricacy of AI architectures calls for automated systems capable of assessing and realigning these models~\cite{peng2023check}. 
Next, the black-box nature of LLMs also highlights the urgency for clarity and explainability in their alignment processes~\cite{zhao2023explainability}.
Lastly, leveraging adversarial attacks as a method to test and refine the alignment of LLMs emerges as an effective approach for ensuring their conformity to human values~\cite{zou2023universal}.

\subsection{Law and Regulatory Compliance}
\label{sec:law}

Given new-emerging ethical challenges posed by LLMs, there is an increasing demand for effective regulation and oversight of LLMs to ensure their safe and responsible use~\cite{cervantes2020artificial}. Regulation refers to the rules, standards, and principles that govern the development, deployment, and use of LLMs, such as laws, policies, guidelines, or codes of conduct~\cite{weaver2018regulation,buiten2019towards,roberts2021chinese}. Oversight refers to the mechanisms, processes, and institutions that monitor, evaluate, and enforce the compliance of LLMs with regulations, such as audits, reviews, certifications, or sanctions~\cite{raji2022outsider}. Regulation and oversight of LLMs aim to protect the rights, interests, and values of the stakeholders involved, such as data owners, users, developers, providers, regulators, and society at large~\cite{mesko2023imperative}.

With that being said, the use of LLMs has not yet been resolved by a consensus or a clear regulation therefore posing ethical and legal challenges.
European Union (EU) has made substantial efforts in the law and regulations on Artificial Intelligence (AI).
In the EU, AI tools, such as LLMs, are subject to the General Data Protection Regulation (GDPR), which regulates the collection, processing, and analysis of personal data, as well as automated decision-making that affects individuals~\cite{regulation2018general}. In this sense, for a company to operate lawfully in the EU regarding the collection and processing of personal data, it must follow the principles and rules laid down in the GDPR.
Furthermore, on May 13, 2022, the French Council presidency circulated an amendment to the draft AI Act~\footnote{https://data.consilium.europa.eu/doc/document/ST-14954-2022-INIT/en/pdf}, on what the text calls “general-purpose AI systems” (GPAIS)~\cite{bertuzzi2023ai, bertuzzi2023ai2}. This novel passage has come to form the nucleus of the direct regulation of LLMs and contains rules on the AI value chain~\cite{bertuzzi2023meps}.

On 30 March 2023, the Italian Data Protection Authority ordered the temporary suspension of the processing of personal data of subjects established on Italian territory by OpenAI LLC, a US company that develops and manages ChatGPT, because the chatbot had failed to comply with the rules set out in GDPR, as well as the Italian Personal Data Protection Code~\cite{patrocinio2023artificial}.
Meanwhile, the EU parliament is continuously working on the EU AI Act, which is poised to be the World's first regulation on AI~\cite{jaeger2024artificial}. This Act envisions a distinct regulatory framework compared to the proposals under consideration in the United Kingdom and categorizes AI systems based on varying risk levels, enabling tailored regulations that correspond to each level of risk~\cite{triguero2024general}.
At the time of writing this manuscript, several other countries are exploring the possibility of limiting or regulating the use of LLMs~\cite{gillespie2023trust,lian2024public}.

\noindent
{\bf Discussion.} Despite the heroic striving of the AI Act to keep up with the accelerating dynamics of AI development, several discussions are also proposed around its practical compliance with LLMs. Hacker et al. argue that this direct regulation is unsatisfactory and could be further enhanced from 1) the definition of GPAIS, 2) the risk management of GPAIS, and 3) the adverse consequences for competition~\cite{hacker2023regulating}. They propose to focus on the deployers and users more and directly apply non-discrimination and data protection law (GDPR compliance) on LLMs. Bommasani et al.~\cite{bommasani2023eu-ai-act} systematically evaluate the compliance with the draft EU AI Act of the foundation model providers like OpenAI and Google. They evaluate the compliance of 10 major foundation model providers (and their flagship models) with the 12 requirements proposed by the EU AI Act and use a scale from 0 (worst) to 4 (best) to rate each provider and model for each requirement. The best possible score for a provider or a model is 48, which indicates full compliance with the AI Act. Their results identify four areas where many organizations receive low scores (usually 0 or 1 out of 4) in terms of compliance with the AI Act: 1) copyrighted data, 2 compute/energy, 3) risk mitigation, and 4) evaluation/testing.
Aside from these general regulations, there are also discussions on challenges of how to regulate LLMs for vertical domains such as medical usage~\cite{minssen2023challenges} and healthcare~\cite{mesko2023imperative}.

\section{Conclusion}
While presenting remarkable opportunities for advancing artificial intelligence (AI) techniques, Large Language Models (LLMs) expose significant ethical challenges that must be meticulously addressed. Exploring the techniques of LLMs within ethical boundaries is a paramount and complicated endeavor, requiring continual innovation in evolving technological capabilities and societal expectations.
In this paper, we survey ethical issues posed by LLMs from longstanding challenges, such as privacy, copyright, and fairness, to new-emerging dilemmas related to truthfulness, social norms, and regulatory compliance. We also discuss the existing approaches that mitigate the potential ethical risks and the corresponding future directions.
Our survey is a stepping stone for researchers to advance LLM techniques under ethical standards, ensuring positive contributions to our society.


\bibliographystyle{plain}
\bibliography{sn-bibliography}

\begin{thebibliography}{100}

\bibitem{abadi2016deep}
Martin Abadi, Andy Chu, Ian Goodfellow, H~Brendan McMahan, Ilya Mironov, Kunal Talwar, and Li~Zhang.
\newblock Deep learning with differential privacy.
\newblock In {\em Proceedings of the 2016 ACM SIGSAC conference on computer and communications security}, pages 308--318, 2016.

\bibitem{abascal2023tmi}
John Abascal, Stanley Wu, Alina Oprea, and Jonathan Ullman.
\newblock Tmi! finetuned models leak private information from their pretraining data.
\newblock {\em arXiv preprint arXiv:2306.01181}, 2023.

\bibitem{abbas2023semdedup}
Amro Abbas, Kushal Tirumala, D{\'a}niel Simig, Surya Ganguli, and Ari~S Morcos.
\newblock Semdedup: Data-efficient learning at web-scale through semantic deduplication.
\newblock {\em arXiv preprint arXiv:2303.09540}, 2023.

\bibitem{abid2021persistent}
Abubakar Abid, Maheen Farooqi, and James Zou.
\newblock Persistent anti-muslim bias in large language models.
\newblock In {\em Proceedings of the 2021 AAAI/ACM Conference on AI, Ethics, and Society}, pages 298--306, 2021.

\bibitem{achiam2023gpt}
Josh Achiam, Steven Adler, Sandhini Agarwal, Lama Ahmad, Ilge Akkaya, Florencia~Leoni Aleman, Diogo Almeida, Janko Altenschmidt, Sam Altman, Shyamal Anadkat, et~al.
\newblock Gpt-4 technical report.
\newblock {\em arXiv preprint arXiv:2303.08774}, 2023.

\bibitem{adi2018turning}
Yossi Adi, Carsten Baum, Moustapha Cisse, Benny Pinkas, and Joseph Keshet.
\newblock Turning your weakness into a strength: Watermarking deep neural networks by backdooring.
\newblock In {\em 27th USENIX Security Symposium (USENIX Security 18)}, pages 1615--1631, 2018.

\bibitem{ahn2021mitigating}
Jaimeen Ahn and Alice Oh.
\newblock Mitigating language-dependent ethnic bias in {BERT}.
\newblock In {\em Proceedings of the 2021 Conference on Empirical Methods in Natural Language Processing}, pages 533--549, Online and Punta Cana, Dominican Republic, November 2021. Association for Computational Linguistics.

\bibitem{Ahvanooey2018ACA}
Milad~Taleby Ahvanooey, Qianmu Li, Hiuk~Jae Shim, and Yanyan Huang.
\newblock A comparative analysis of information hiding techniques for copyright protection of text documents.
\newblock {\em Secur. Commun. Networks}, 2018:5325040:1--5325040:22, 2018.

\bibitem{Ahvanooey2016AnIT}
Milad~Taleby Ahvanooey, Hassan~Dana Mazraeh, and Seyed~Hashem Tabasi.
\newblock An innovative technique for web text watermarking (aitw).
\newblock {\em Information Security Journal: A Global Perspective}, 25:191 -- 196, 2016.

\bibitem{anil2021large}
Rohan Anil, Badih Ghazi, Vineet Gupta, Ravi Kumar, and Pasin Manurangsi.
\newblock Large-scale differentially private bert.
\newblock {\em arXiv preprint arXiv:2108.01624}, 2021.

\bibitem{askell2021general}
Amanda Askell, Yuntao Bai, Anna Chen, Dawn Drain, Deep Ganguli, Tom Henighan, Andy Jones, Nicholas Joseph, Ben Mann, Nova DasSarma, et~al.
\newblock A general language assistant as a laboratory for alignment.
\newblock {\em arXiv preprint arXiv:2112.00861}, 2021.

\bibitem{Atallah2002watermark}
Mikhail~J. Atallah, Victor Raskin, Christian~F. Hempelmann, Mercan Karahan, Radu Sion, Umut Topkara, and Katrina~E. Triezenberg.
\newblock Natural language watermarking and tamperproofing.
\newblock In {\em International Workshop on Information Hiding}, pages 196--212, 2002.

\bibitem{attanasio2022entropy}
Giuseppe Attanasio, Debora Nozza, Dirk Hovy, and Elena Baralis.
\newblock Entropy-based attention regularization frees unintended bias mitigation from lists.
\newblock In {\em Findings of the Association for Computational Linguistics: ACL 2022}, pages 1105--1119, Dublin, Ireland, May 2022. Association for Computational Linguistics.

\bibitem{azamfirei2023large}
Razvan Azamfirei, Sapna~R Kudchadkar, and James Fackler.
\newblock Large language models and the perils of their hallucinations.
\newblock {\em Critical Care}, 27(1):120, 2023.

\bibitem{baek2023knowledge}
Jinheon Baek, Alham~Fikri Aji, and Amir Saffari.
\newblock Knowledge-augmented language model prompting for zero-shot knowledge graph question answering.
\newblock {\em arXiv:2306.04136}, 2023.

\bibitem{bai2023badclip}
Jiawang Bai, Kuofeng Gao, Shaobo Min, Shu-Tao Xia, Zhifeng Li, and Wei Liu.
\newblock Badclip: Trigger-aware prompt learning for backdoor attacks on clip.
\newblock In {\em CVPR}, 2024.

\bibitem{bai2022training}
Yuntao Bai, Andy Jones, Kamal Ndousse, Amanda Askell, Anna Chen, Nova DasSarma, Dawn Drain, Stanislav Fort, Deep Ganguli, Tom Henighan, et~al.
\newblock Training a helpful and harmless assistant with reinforcement learning from human feedback.
\newblock {\em arXiv preprint arXiv:2204.05862}, 2022.

\bibitem{bai2022constitutional}
Yuntao Bai, Saurav Kadavath, Sandipan Kundu, Amanda Askell, Jackson Kernion, Andy Jones, Anna Chen, Anna Goldie, Azalia Mirhoseini, Cameron McKinnon, et~al.
\newblock Constitutional ai: Harmlessness from ai feedback.
\newblock {\em arXiv preprint arXiv:2212.08073}, 2022.

\bibitem{balunovic2022lamp}
Mislav Balunovic, Dimitar Dimitrov, Nikola Jovanovi{\'c}, and Martin Vechev.
\newblock Lamp: Extracting text from gradients with language model priors.
\newblock {\em Advances in Neural Information Processing Systems}, 35:7641--7654, 2022.

\bibitem{fairmlbook2019}
Solon Barocas, Moritz Hardt, and Arvind Narayanan.
\newblock {\em Fairness and Machine Learning: Limitations and Opportunities}.
\newblock fairmlbook.org, 2019.
\newblock \url{http://www.fairmlbook.org}.

\bibitem{bender2021dangers}
Emily~M Bender, Timnit Gebru, Angelina McMillan-Major, and Shmargaret Shmitchell.
\newblock On the dangers of stochastic parrots: Can language models be too big?
\newblock In {\em Proceedings of the 2021 ACM conference on fairness, accountability, and transparency}, pages 610--623, 2021.

\bibitem{Bender2000ApplicationsFD}
Walter Bender, William Butera, Daniel~F. Gruhl, Raymond Hwang, Fernando~J. Paiz, and Sofya Pogreb.
\newblock Applications for data hiding.
\newblock {\em IBM Syst. J.}, 39:547--568, 2000.

\bibitem{Bender1995TechniquesFD}
Walter Bender, Daniel~F. Gruhl, Norishige Morimoto, and Anthony Lu.
\newblock Techniques for data hiding.
\newblock In {\em Electronic imaging}, 1995.

\bibitem{Bergmair2004TowardsLS}
Richard Bergmair.
\newblock Towards linguistic steganography: A systematic investigation of approaches, systems, and issues.
\newblock 2004.

\bibitem{berk2017convex}
Richard Berk, Hoda Heidari, Shahin Jabbari, Matthew Joseph, Michael Kearns, Jamie Morgenstern, Seth Neel, and Aaron Roth.
\newblock A convex framework for fair regression.
\newblock {\em arXiv preprint arXiv:1706.02409}, 2017.

\bibitem{bertuzzi2023ai}
L~Bertuzzi.
\newblock Ai act: Eu parliament’s crunch time on high-risk categorisation, prohibited practices, 2023.

\bibitem{bertuzzi2023ai2}
L~Bertuzzi.
\newblock Ai act: Meps close in on rules for general purpose ai, foundation models, 2023.

\bibitem{bertuzzi2023meps}
L~Bertuzzi.
\newblock Meps seal the deal on artificial intelligence act, 2023.

\bibitem{beukeboom2019stereotypes}
Camiel~J Beukeboom and Christian Burgers.
\newblock How stereotypes are shared through language: a review and introduction of the aocial categories and stereotypes communication (scsc) framework.
\newblock {\em Review of Communication Research}, 7:1--37, 2019.

\bibitem{biderman2023pythia}
Stella Biderman, Hailey Schoelkopf, Quentin~Gregory Anthony, Herbie Bradley, Kyle O’Brien, Eric Hallahan, Mohammad~Aflah Khan, Shivanshu Purohit, USVSN~Sai Prashanth, Edward Raff, et~al.
\newblock Pythia: A suite for analyzing large language models across training and scaling.
\newblock In {\em International Conference on Machine Learning}, pages 2397--2430. PMLR, 2023.

\bibitem{bishop2010relationships}
Matt Bishop, Justin Cummins, Sean Peisert, Anhad Singh, Bhume Bhumiratana, Deborah Agarwal, Deborah Frincke, and Michael Hogarth.
\newblock Relationships and data sanitization: A study in scarlet.
\newblock In {\em Proceedings of the 2010 New Security Paradigms Workshop}, pages 151--164, 2010.

\bibitem{blodgett2020language}
Su~Lin Blodgett, Solon Barocas, Hal Daum{\'e}~III, and Hanna Wallach.
\newblock Language (technology) is power: A critical survey of {``}bias{''} in {NLP}.
\newblock In {\em Proceedings of the 58th Annual Meeting of the Association for Computational Linguistics}, pages 5454--5476, Online, July 2020. Association for Computational Linguistics.

\bibitem{blodgett2017racial}
Su~Lin Blodgett and Brendan O'Connor.
\newblock Racial disparity in natural language processing: A case study of social media african-american english.
\newblock {\em arXiv preprint arXiv:1707.00061}, 2017.

\bibitem{bommasani2023eu-ai-act}
Rishi Bommasani, Kevin Klyman, Daniel Zhang, and Percy Liang.
\newblock Do foundation model providers comply with the eu ai act?, 2023.

\bibitem{bommasani2023holistic}
Rishi Bommasani, Percy Liang, and Tony Lee.
\newblock Holistic evaluation of language models.
\newblock {\em Annals of the New York Academy of Sciences}, 2023.

\bibitem{Brassil1999CopyrightPF}
Jack Brassil, Steven~H. Low, and Nicholas~F. Maxemchuk.
\newblock Copyright protection for the electronic distribution of text documents.
\newblock {\em Proc. IEEE}, 87:1181--1196, 1999.

\bibitem{Brassil1995discopy}
J.T. Brassil, S.~Low, N.F. Maxemchuk, and L.~O'Gorman.
\newblock Electronic marking and identification techniques to discourage document copying.
\newblock {\em IEEE Journal on Selected Areas in Communications}, 13(8):1495--1504, 1995.

\bibitem{brown2020language}
Tom Brown, Benjamin Mann, Nick Ryder, Melanie Subbiah, Jared~D Kaplan, Prafulla Dhariwal, Arvind Neelakantan, Pranav Shyam, Girish Sastry, Amanda Askell, et~al.
\newblock Language models are few-shot learners.
\newblock {\em Advances in neural information processing systems}, 33:1877--1901, 2020.

\bibitem{buiten2019towards}
Miriam~C Buiten.
\newblock Towards intelligent regulation of artificial intelligence.
\newblock {\em European Journal of Risk Regulation}, 10(1):41--59, 2019.

\bibitem{caliskan2017semantics}
Aylin Caliskan, Joanna~J. Bryson, and Arvind Narayanan.
\newblock Semantics derived automatically from language corpora contain human-like biases.
\newblock {\em Science}, 356(6334):183--186, 2017.

\bibitem{cao2023assessing}
Yong Cao, Li~Zhou, Seolhwa Lee, Laura Cabello, Min Chen, and Daniel Hershcovich.
\newblock Assessing cross-cultural alignment between chatgpt and human societies: An empirical study.
\newblock {\em arXiv preprint arXiv:2303.17466}, 2023.

\bibitem{carlini2022membership}
Nicholas Carlini, Steve Chien, Milad Nasr, Shuang Song, Andreas Terzis, and Florian Tramer.
\newblock Membership inference attacks from first principles.
\newblock In {\em 2022 IEEE Symposium on Security and Privacy (SP)}, pages 1897--1914. IEEE, 2022.

\bibitem{carlini2021extracting}
Nicholas Carlini, Florian Tramer, Eric Wallace, Matthew Jagielski, Ariel Herbert-Voss, Katherine Lee, Adam Roberts, Tom Brown, Dawn Song, Ulfar Erlingsson, et~al.
\newblock Extracting training data from large language models.
\newblock In {\em 30th USENIX Security Symposium (USENIX Security 21)}, pages 2633--2650, 2021.

\bibitem{carlini2023extracting}
Nicolas Carlini, Jamie Hayes, Milad Nasr, Matthew Jagielski, Vikash Sehwag, Florian Tramer, Borja Balle, Daphne Ippolito, and Eric Wallace.
\newblock Extracting training data from diffusion models.
\newblock In {\em 32nd USENIX Security Symposium (USENIX Security 23)}, pages 5253--5270, 2023.

\bibitem{cervantes2020artificial}
Jos{\'e}-Antonio Cervantes, Sonia L{\'o}pez, Luis-Felipe Rodr{\'\i}guez, Salvador Cervantes, Francisco Cervantes, and F{\'e}lix Ramos.
\newblock Artificial moral agents: A survey of the current status.
\newblock {\em Science and engineering ethics}, 26:501--532, 2020.

\bibitem{chang2023survey}
Yupeng Chang, Xu~Wang, Jindong Wang, Yuan Wu, Kaijie Zhu, Hao Chen, Linyi Yang, Xiaoyuan Yi, Cunxiang Wang, Yidong Wang, et~al.
\newblock A survey on evaluation of large language models.
\newblock {\em arXiv preprint arXiv:2307.03109}, 2023.

\bibitem{chen2023semi}
Changyu Chen, Xiting Wang, Yiqiao Jin, Victor~Ye Dong, Li~Dong, Jie Cao, Yi~Liu, and Rui Yan.
\newblock Semi-offline reinforcement learning for optimized text generation.
\newblock In {\em ICML}, 2023.

\bibitem{chen2023rna}
Dake Chen, Yuke Zhang, Souvik Kundu, Chenghao Li, and Peter~A Beerel.
\newblock Rna-vit: Reduced-dimension approximate normalized attention vision transformers for latency efficient private inference.
\newblock In {\em 2023 IEEE/ACM International Conference on Computer Aided Design (ICCAD)}, pages 1--9. IEEE, 2023.

\bibitem{chen2022x}
Tianyu Chen, Hangbo Bao, Shaohan Huang, Li~Dong, Binxing Jiao, Daxin Jiang, Haoyi Zhou, Jianxin Li, and Furu Wei.
\newblock The-x: Privacy-preserving transformer inference with homomorphic encryption.
\newblock {\em arXiv preprint arXiv:2206.00216}, 2022.

\bibitem{chen2017targeted}
Xinyun Chen, Chang Liu, Bo~Li, Kimberly Lu, and Dawn Song.
\newblock Targeted backdoor attacks on deep learning systems using data poisoning.
\newblock {\em arXiv preprint arXiv:1712.05526}, 2017.

\bibitem{chen2023exploring}
Yi~Chen, Rui Wang, Haiyun Jiang, Shuming Shi, and Ruifeng Xu.
\newblock Exploring the use of large language models for reference-free text quality evaluation: A preliminary empirical study.
\newblock {\em arXiv preprint arXiv:2304.00723}, 2023.

\bibitem{cheng2021socially}
Lu~Cheng, Kush~R Varshney, and Huan Liu.
\newblock Socially responsible ai algorithms: Issues, purposes, and challenges.
\newblock {\em Journal of Artificial Intelligence Research}, 71:1137--1181, 2021.

\bibitem{chouldechova2017fair}
Alexandra Chouldechova.
\newblock Fair prediction with disparate impact: A study of bias in recidivism prediction instruments.
\newblock {\em Big data}, 5(2):153--163, 2017.

\bibitem{Christ2023UndetectableWF}
Miranda Christ, Sam Gunn, and Or~Zamir.
\newblock Undetectable watermarks for language models.
\newblock {\em ArXiv}, abs/2306.09194, 2023.

\bibitem{christiano2017deep}
Paul~F Christiano, Jan Leike, Tom Brown, Miljan Martic, Shane Legg, and Dario Amodei.
\newblock Deep reinforcement learning from human preferences.
\newblock {\em Advances in neural information processing systems}, 30, 2017.

\bibitem{chu2022panning}
Hong-Min Chu, Jonas Geiping, Liam~H Fowl, Micah Goldblum, and Tom Goldstein.
\newblock Panning for gold in federated learning: Targeted text extraction under arbitrarily large-scale aggregation.
\newblock In {\em The Eleventh International Conference on Learning Representations}, 2022.

\bibitem{chung2022scaling}
Hyung~Won Chung, Le~Hou, Shayne Longpre, Barret Zoph, Yi~Tay, William Fedus, Eric Li, Xuezhi Wang, Mostafa Dehghani, Siddhartha Brahma, et~al.
\newblock Scaling instruction-finetuned language models.
\newblock {\em arXiv preprint arXiv:2210.11416}, 2022.

\bibitem{chung2023increasing}
John Chung, Ece Kamar, and Saleema Amershi.
\newblock Increasing diversity while maintaining accuracy: Text data generation with large language models and human interventions.
\newblock In {\em Proceedings of the 61st Annual Meeting of the Association for Computational Linguistics (Volume 1: Long Papers)}, pages 575--593, Toronto, Canada, July 2023. Association for Computational Linguistics.

\bibitem{CPSumathi2013ASO}
C.P.Sumathi, T.Santanam, Graduate~School of~Science, Sdnb Vaishnav College~For Women, Chennai, Indian~Institute of~Science, DG~Vaishnav College~For Men, and India.
\newblock A study of various steganographic techniques used for information hiding.
\newblock {\em ArXiv}, abs/1401.5561, 2013.

\bibitem{cramer2015secure}
Ronald Cramer, Ivan~Bjerre Damg{\aa}rd, et~al.
\newblock {\em Secure multiparty computation}.
\newblock Cambridge University Press, 2015.

\bibitem{crawford2017trouble}
Kate Crawford.
\newblock The trouble with bias, 2017.
\newblock Keynote at NeurIPS.

\bibitem{cui2022unified}
Ganqu Cui, Lifan Yuan, Bingxiang He, Yangyi Chen, Zhiyuan Liu, and Maosong Sun.
\newblock A unified evaluation of textual backdoor learning: Frameworks and benchmarks.
\newblock {\em Advances in Neural Information Processing Systems}, 35:5009--5023, 2022.

\bibitem{d2017conscientious}
Brian d'Alessandro, Cathy O'Neil, and Tom LaGatta.
\newblock Conscientious classification: A data scientist's guide to discrimination-aware classification.
\newblock {\em Big data}, 5(2):120--134, 2017.

\bibitem{delobelle2022fairdistillation}
Pieter Delobelle and Bettina Berendt.
\newblock Fairdistillation: mitigating stereotyping in language models.
\newblock In {\em Joint European Conference on Machine Learning and Knowledge Discovery in Databases}, pages 638--654. Springer, 2022.

\bibitem{dev2020measuring}
Sunipa Dev, Tao Li, Jeff~M Phillips, and Vivek Srikumar.
\newblock On measuring and mitigating biased inferences of word embeddings.
\newblock In {\em Proceedings of the AAAI Conference on Artificial Intelligence}, volume~34, pages 7659--7666, 2020.

\bibitem{dhamala2021bold}
Jwala Dhamala, Tony Sun, Varun Kumar, Satyapriya Krishna, Yada Pruksachatkun, Kai-Wei Chang, and Rahul Gupta.
\newblock Bold: Dataset and metrics for measuring biases in open-ended language generation.
\newblock In {\em Proceedings of the 2021 ACM Conference on Fairness, Accountability, and Transparency}, FAccT '21, page 862–872, New York, NY, USA, 2021. Association for Computing Machinery.

\bibitem{dhingra2023queer}
Harnoor Dhingra, Preetiha Jayashanker, Sayali Moghe, and Emma Strubell.
\newblock Queer people are people first: Deconstructing sexual identity stereotypes in large language models.
\newblock {\em arXiv preprint arXiv:2307.00101}, 2023.

\bibitem{dolci2023improving}
Tommaso Dolci, Fabio Azzalini, and Mara Tanelli.
\newblock Improving gender-related fairness in sentence encoders: A semantics-based approach.
\newblock {\em Data Science and Engineering}, pages 1--19, 2023.

\bibitem{dong2023puma}
Ye~Dong, Wen-jie Lu, Yancheng Zheng, Haoqi Wu, Derun Zhao, Jin Tan, Zhicong Huang, Cheng Hong, Tao Wei, and Wenguang Cheng.
\newblock Puma: Secure inference of llama-7b in five minutes.
\newblock {\em arXiv preprint arXiv:2307.12533}, 2023.

\bibitem{du2023dp}
Minxin Du, Xiang Yue, Sherman~SM Chow, Tianhao Wang, Chenyu Huang, and Huan Sun.
\newblock Dp-forward: Fine-tuning and inference on language models with differential privacy in forward pass.
\newblock In {\em Proceedings of the 2023 ACM SIGSAC Conference on Computer and Communications Security}, pages 2665--2679, 2023.

\bibitem{dwork2012fairness}
Cynthia Dwork, Moritz Hardt, Toniann Pitassi, Omer Reingold, and Richard Zemel.
\newblock Fairness through awareness.
\newblock In {\em Proceedings of the 3rd Innovations in Theoretical Computer Science Conference}, ITCS '12, page 214–226, New York, NY, USA, 2012. Association for Computing Machinery.

\bibitem{dwork2014algorithmic}
Cynthia Dwork and Aaron Roth.
\newblock The algorithmic foundations of differential privacy.
\newblock {\em Foundations and Trends{\textregistered} in Theoretical Computer Science}, 9(3--4):211--407, 2014.

\bibitem{fereidooni2021safelearn}
Hossein Fereidooni, Samuel Marchal, Markus Miettinen, Azalia Mirhoseini, Helen M{\"o}llering, Thien~Duc Nguyen, Phillip Rieger, Ahmad-Reza Sadeghi, Thomas Schneider, Hossein Yalame, et~al.
\newblock Safelearn: Secure aggregation for private federated learning.
\newblock In {\em 2021 IEEE Security and Privacy Workshops (SPW)}, pages 56--62. IEEE, 2021.

\bibitem{ferrara2023should}
Emilio Ferrara.
\newblock Should chatgpt be biased? challenges and risks of bias in large language models.
\newblock {\em arXiv preprint arXiv:2304.03738}, 2023.

\bibitem{fowl2022decepticons}
Liam Fowl, Jonas Geiping, Steven Reich, Yuxin Wen, Wojtek Czaja, Micah Goldblum, and Tom Goldstein.
\newblock Decepticons: Corrupted transformers breach privacy in federated learning for language models.
\newblock {\em arXiv preprint arXiv:2201.12675}, 2022.

\bibitem{Fu2023WatermarkingCT}
Yu~Fu, Deyi Xiong, and Yue Dong.
\newblock Watermarking conditional text generation for ai detection: Unveiling challenges and a semantic-aware watermark remedy.
\newblock {\em ArXiv}, abs/2307.13808, 2023.

\bibitem{gaci2022debiasing}
Yacine Gaci, Boualem Benattallah, Fabio Casati, and Khalid Benabdeslem.
\newblock {Debiasing Pretrained Text Encoders by Paying Attention to Paying Attention}.
\newblock In {\em {2022 Conference on Empirical Methods in Natural Language Processing}}, Proceedings of the 2022 Conference on Empirical Methods in Natural Language Processing, pages 9582--9602, Abu Dhabi, United Arab Emirates, December 2022. Association for Computational Linguistics, {Association for Computational Linguistics}.

\bibitem{gallegos2023bias}
Isabel~O Gallegos, Ryan~A Rossi, Joe Barrow, Md~Mehrab Tanjim, Sungchul Kim, Franck Dernoncourt, Tong Yu, Ruiyi Zhang, and Nesreen~K Ahmed.
\newblock Bias and fairness in large language models: A survey.
\newblock {\em arXiv preprint arXiv:2309.00770}, 2023.

\bibitem{gao2023imperceptible}
Kuofeng Gao, Jiawang Bai, Baoyuan Wu, Mengxi Ya, and Shu-Tao Xia.
\newblock Imperceptible and robust backdoor attack in 3d point cloud.
\newblock {\em IEEE Transactions on Information Forensics and Security}, 19:1267--1282, 2023.

\bibitem{gao2024inducing}
Kuofeng Gao, Yang Bai, Jindong Gu, Shu-Tao Xia, Philip Torr, Zhifeng Li, and Wei Liu.
\newblock Inducing high energy-latency of large vision-language models with verbose images.
\newblock In {\em International Conference on Learning Representations}, 2024.

\bibitem{gao2023backdoor}
Kuofeng Gao, Yang Bai, Jindong Gu, Yong Yang, and Shu-Tao Xia.
\newblock Backdoor defense via adaptively splitting poisoned dataset.
\newblock In {\em Proceedings of the IEEE/CVF Conference on Computer Vision and Pattern Recognition}, pages 4005--4014, 2023.

\bibitem{gao2024energy}
Kuofeng Gao, Jindong Gu, Yang Bai, Shu-Tao Xia, Philip Torr, Wei Liu, and Zhifeng Li.
\newblock Energy-latency manipulation of multi-modal large language models via verbose samples.
\newblock {\em arXiv preprint arXiv:2404.16557}, 2024.

\bibitem{gehrmann2019gltr}
Sebastian Gehrmann, Hendrik Strobelt, and Alexander~M Rush.
\newblock Gltr: Statistical detection and visualization of generated text.
\newblock {\em arXiv preprint arXiv:1906.04043}, 2019.

\bibitem{geng2024large}
Mingmeng Geng, Sihong He, and Roberto Trotta.
\newblock Are large language models chameleons?
\newblock {\em arXiv preprint arXiv:2405.19323}, 2024.

\bibitem{ghanbarzadeh2023gender}
Somayeh Ghanbarzadeh, Yan Huang, Hamid Palangi, Radames Cruz~Moreno, and Hamed Khanpour.
\newblock Gender-tuning: Empowering fine-tuning for debiasing pre-trained language models.
\newblock In {\em Findings of the Association for Computational Linguistics: ACL 2023}, pages 5448--5458, Toronto, Canada, July 2023. Association for Computational Linguistics.

\bibitem{gillespie2023trust}
Nicole Gillespie, Steven Lockey, Caitlin Curtis, Javad Pool, and Ali Akbari.
\newblock Trust in artificial intelligence: A global study.
\newblock 2023.

\bibitem{goldreich1998secure}
Oded Goldreich.
\newblock Secure multi-party computation.
\newblock {\em Manuscript. Preliminary version}, 78(110), 1998.

\bibitem{gu2022watermarking}
Chenxi Gu, Chengsong Huang, Xiaoqing Zheng, Kai-Wei Chang, and Cho-Jui Hsieh.
\newblock Watermarking pre-trained language models with backdooring.
\newblock {\em arXiv preprint arXiv:2210.07543}, 2022.

\bibitem{gu2019badnets}
Tianyu Gu, Kang Liu, Brendan Dolan-Gavitt, and Siddharth Garg.
\newblock Badnets: Evaluating backdooring attacks on deep neural networks.
\newblock {\em IEEE Access}, 7:47230--47244, 2019.

\bibitem{gunasekar2023textbooks}
Suriya Gunasekar, Yi~Zhang, Jyoti Aneja, Caio C{\'e}sar~Teodoro Mendes, Allie Del~Giorno, Sivakanth Gopi, Mojan Javaheripi, Piero Kauffmann, Gustavo de~Rosa, Olli Saarikivi, et~al.
\newblock Textbooks are all you need.
\newblock {\em arXiv preprint arXiv:2306.11644}, 2023.

\bibitem{guo2021detecting}
Wei Guo and Aylin Caliskan.
\newblock Detecting emergent intersectional biases: Contextualized word embeddings contain a distribution of human-like biases.
\newblock In {\em Proceedings of the 2021 AAAI/ACM Conference on AI, Ethics, and Society}, pages 122--133, 2021.

\bibitem{gupta2023sigma}
Kanav Gupta, Neha Jawalkar, Ananta Mukherjee, Nishanth Chandran, Divya Gupta, Ashish Panwar, and Rahul Sharma.
\newblock Sigma: secure gpt inference with function secret sharing.
\newblock {\em Cryptology ePrint Archive}, 2023.

\bibitem{gupta2022recovering}
Samyak Gupta, Yangsibo Huang, Zexuan Zhong, Tianyu Gao, Kai Li, and Danqi Chen.
\newblock Recovering private text in federated learning of language models.
\newblock {\em Advances in Neural Information Processing Systems}, 35:8130--8143, 2022.

\bibitem{gupta2022mitigating}
Umang Gupta, Jwala Dhamala, Varun Kumar, Apurv Verma, Yada Pruksachatkun, Satyapriya Krishna, Rahul Gupta, Kai-Wei Chang, Greg Ver~Steeg, and Aram Galstyan.
\newblock Mitigating gender bias in distilled language models via counterfactual role reversal.
\newblock In {\em Findings of the Association for Computational Linguistics: ACL 2022}, pages 658--678, Dublin, Ireland, May 2022. Association for Computational Linguistics.

\bibitem{hacker2023regulating}
Philipp Hacker, Andreas Engel, and Marco Mauer.
\newblock Regulating chatgpt and other large generative ai models.
\newblock In {\em Proceedings of the 2023 ACM Conference on Fairness, Accountability, and Transparency}, pages 1112--1123, 2023.

\bibitem{hallinan2022detoxifying}
Skyler Hallinan, Alisa Liu, Yejin Choi, and Maarten Sap.
\newblock Detoxifying text with marco: Controllable revision with experts and anti-experts.
\newblock {\em arXiv preprint arXiv:2212.10543}, 2022.

\bibitem{han2021diverse}
Xudong Han, Timothy Baldwin, and Trevor Cohn.
\newblock Diverse adversaries for mitigating bias in training.
\newblock In {\em Proceedings of the 16th Conference of the European Chapter of the Association for Computational Linguistics: Main Volume}, pages 2760--2765, Online, April 2021. Association for Computational Linguistics.

\bibitem{han2022balancing}
Xudong Han, Timothy Baldwin, and Trevor Cohn.
\newblock Balancing out bias: Achieving fairness through balanced training.
\newblock In {\em Proceedings of the 2022 Conference on Empirical Methods in Natural Language Processing}, pages 11335--11350, Abu Dhabi, United Arab Emirates, December 2022. Association for Computational Linguistics.

\bibitem{hao2022iron}
Meng Hao, Hongwei Li, Hanxiao Chen, Pengzhi Xing, Guowen Xu, and Tianwei Zhang.
\newblock Iron: Private inference on transformers.
\newblock {\em Advances in Neural Information Processing Systems}, 35:15718--15731, 2022.

\bibitem{hardt2016equality}
Moritz Hardt, Eric Price, and Nati Srebro.
\newblock Equality of opportunity in supervised learning.
\newblock {\em Advances in neural information processing systems}, 29, 2016.

\bibitem{hauzenberger2023modular}
Lukas Hauzenberger, Shahed Masoudian, Deepak Kumar, Markus Schedl, and Navid Rekabsaz.
\newblock Modular and on-demand bias mitigation with attribute-removal subnetworks.
\newblock In {\em Findings of the Association for Computational Linguistics: ACL 2023}, pages 6192--6214, 2023.

\bibitem{he2023robust_a}
Sihong He, Shuo Han, and Fei Miao.
\newblock Robust electric vehicle balancing of autonomous mobility-on-demand system: A multi-agent reinforcement learning approach.
\newblock In {\em 2023 IEEE/RSJ International Conference on Intelligent Robots and Systems (IROS)}, pages 5471--5478. IEEE, 2023.

\bibitem{he2020data}
Sihong He, Lynn Pepin, Guang Wang, Desheng Zhang, and Fei Miao.
\newblock Data-driven distributionally robust electric vehicle balancing for mobility-on-demand systems under demand and supply uncertainties.
\newblock In {\em 2020 IEEE/RSJ International Conference on Intelligent Robots and Systems (IROS)}, pages 2165--2172. IEEE, 2020.

\bibitem{he2022robust}
Sihong He, Yue Wang, Shuo Han, Shaofeng Zou, and Fei Miao.
\newblock A robust and constrained multi-agent reinforcement learning framework for electric vehicle amod systems.
\newblock {\em arXiv preprint arXiv:2209.08230}, 2022.

\bibitem{he2023robust_b}
Sihong He, Yue Wang, Shuo Han, Shaofeng Zou, and Fei Miao.
\newblock A robust and constrained multi-agent reinforcement learning electric vehicle rebalancing method in amod systems.
\newblock In {\em 2023 IEEE/RSJ International Conference on Intelligent Robots and Systems (IROS)}, pages 5637--5644. IEEE, 2023.

\bibitem{he2023data}
Sihong He, Zhili Zhang, Shuo Han, Lynn Pepin, Guang Wang, Desheng Zhang, John~A Stankovic, and Fei Miao.
\newblock Data-driven distributionally robust electric vehicle balancing for autonomous mobility-on-demand systems under demand and supply uncertainties.
\newblock {\em IEEE Transactions on Intelligent Transportation Systems}, 2023.

\bibitem{he2022protecting}
Xuanli He, Qiongkai Xu, Lingjuan Lyu, Fangzhao Wu, and Chenguang Wang.
\newblock Protecting intellectual property of language generation apis with lexical watermark.
\newblock In {\em Proceedings of the AAAI Conference on Artificial Intelligence}, volume~36, pages 10758--10766, 2022.

\bibitem{he2021detect}
Zexue He, Bodhisattwa~Prasad Majumder, and Julian McAuley.
\newblock Detect and perturb: Neutral rewriting of biased and sensitive text via gradient-based decoding.
\newblock In {\em Findings of the Association for Computational Linguistics: EMNLP 2021}, pages 4173--4181, Punta Cana, Dominican Republic, November 2021. Association for Computational Linguistics.

\bibitem{he2022controlling}
Zexue He, Yu~Wang, Julian McAuley, and Bodhisattwa~Prasad Majumder.
\newblock Controlling bias exposure for fair interpretable predictions.
\newblock In {\em Findings of the Association for Computational Linguistics: EMNLP 2022}, pages 5854--5866, Abu Dhabi, United Arab Emirates, December 2022. Association for Computational Linguistics.

\bibitem{hinton2015distilling}
Geoffrey Hinton, Oriol Vinyals, and Jeff Dean.
\newblock Distilling the knowledge in a neural network.
\newblock In {\em NeurIPS Workshop}, 2014.

\bibitem{Hisamoto_2020}
Sorami Hisamoto, Matt Post, and Kevin Duh.
\newblock Membership inference attacks on sequence-to-sequence models: Is my data in your machine translation system?
\newblock {\em Transactions of the Association for Computational Linguistics}, 8:49–63, December 2020.

\bibitem{hoory2021learning}
Shlomo Hoory, Amir Feder, Avichai Tendler, Sofia Erell, Alon Peled-Cohen, Itay Laish, Hootan Nakhost, Uri Stemmer, Ayelet Benjamini, Avinatan Hassidim, et~al.
\newblock Learning and evaluating a differentially private pre-trained language model.
\newblock In {\em Findings of the Association for Computational Linguistics: EMNLP 2021}, pages 1178--1189, 2021.

\bibitem{Hou2023SemStampAS}
Abe~Bohan Hou, Jingyu Zhang, Tianxing He, Yichen Wang, Yung-Sung Chuang, Hongwei Wang, Lingfeng Shen, Benjamin~Van Durme, Daniel Khashabi, and Yulia Tsvetkov.
\newblock Semstamp: A semantic watermark with paraphrastic robustness for text generation.
\newblock {\em ArXiv}, abs/2310.03991, 2023.

\bibitem{hou2024k}
Abe~Bohan Hou, Jingyu Zhang, Yichen Wang, Daniel Khashabi, and Tianxing He.
\newblock k-semstamp: A clustering-based semantic watermark for detection of machine-generated text.
\newblock {\em arXiv preprint arXiv:2402.11399}, 2024.

\bibitem{hou2023privately}
Charlie Hou, Hongyuan Zhan, Akshat Shrivastava, Sid Wang, Sasha Livshits, Giulia Fanti, and Daniel Lazar.
\newblock Privately customizing prefinetuning to better match user data in federated learning.
\newblock {\em arXiv preprint arXiv:2302.09042}, 2023.

\bibitem{hou2023ciphergpt}
Xiaoyang Hou, Jian Liu, Jingyu Li, Yuhan Li, Wen-jie Lu, Cheng Hong, and Kui Ren.
\newblock Ciphergpt: Secure two-party gpt inference.
\newblock {\em Cryptology ePrint Archive}, 2023.

\bibitem{houlsby2019parameter}
Neil Houlsby, Andrei Giurgiu, Stanislaw Jastrzebski, Bruna Morrone, Quentin De~Laroussilhe, Andrea Gesmundo, Mona Attariyan, and Sylvain Gelly.
\newblock Parameter-efficient transfer learning for nlp.
\newblock In {\em International Conference on Machine Learning}, pages 2790--2799. PMLR, 2019.

\bibitem{hu2023artificial}
Tiechuan Hu, Wenbo Zhu, and Yuqi Yan.
\newblock Artificial intelligence aspect of transportation analysis using large scale systems.
\newblock In {\em 2023 6th Artificial Intelligence and Cloud Computing Conference (AICCC)}, pages 54--59, 2023.

\bibitem{huang2022towards}
Jie Huang and Kevin Chen-Chuan Chang.
\newblock Towards reasoning in large language models: A survey.
\newblock {\em arXiv preprint arXiv:2212.10403}, 2022.

\bibitem{huang2023survey}
Lei Huang, Weijiang Yu, Weitao Ma, Weihong Zhong, Zhangyin Feng, Haotian Wang, Qianglong Chen, Weihua Peng, Xiaocheng Feng, Bing Qin, et~al.
\newblock A survey on hallucination in large language models: Principles, taxonomy, challenges, and open questions.
\newblock {\em arXiv preprint arXiv:2311.05232}, 2023.

\bibitem{huang2020reducing}
Po-Sen Huang, Huan Zhang, Ray Jiang, Robert Stanforth, Johannes Welbl, Jack Rae, Vishal Maini, Dani Yogatama, and Pushmeet Kohli.
\newblock Reducing sentiment bias in language models via counterfactual evaluation.
\newblock In {\em Findings of the Association for Computational Linguistics: EMNLP 2020}, pages 65--83, Online, November 2020. Association for Computational Linguistics.

\bibitem{huang2023trustgpt}
Yue Huang, Qihui Zhang, Lichao Sun, et~al.
\newblock Trustgpt: A benchmark for trustworthy and responsible large language models.
\newblock {\em arXiv preprint arXiv:2306.11507}, 2023.

\bibitem{ishibashi2023knowledge}
Yoichi Ishibashi and Hidetoshi Shimodaira.
\newblock Knowledge sanitization of large language models.
\newblock {\em arXiv preprint arXiv:2309.11852}, 2023.

\bibitem{iskander2023shielded}
Shadi Iskander, Kira Radinsky, and Yonatan Belinkov.
\newblock Shielded representations: Protecting sensitive attributes through iterative gradient-based projection.
\newblock In {\em Findings of the Association for Computational Linguistics: ACL 2023}, pages 5961--5977, Toronto, Canada, July 2023. Association for Computational Linguistics.

\bibitem{jaeger2024artificial}
Lars Jaeger and Michel Dacorogna.
\newblock Artificial intelligence from its origins via today to the future: Significant progress in understanding, replicating, and changing us humans or solely technological advances contained to optimising certain processes?
\newblock In {\em Where Is Science Leading Us? And What Can We Do to Steer It?}, pages 207--235. Springer, 2024.

\bibitem{jain2021generating}
Nishtha Jain, Maja Popovi{\'c}, Declan Groves, and Eva Vanmassenhove.
\newblock Generating gender augmented data for {NLP}.
\newblock In {\em Proceedings of the 3rd Workshop on Gender Bias in Natural Language Processing}, pages 93--102, Online, August 2021. Association for Computational Linguistics.

\bibitem{jalil2010word}
Zunera Jalil, Anwar~M Mirza, and Hajira Jabeen.
\newblock Word length based zero-watermarking algorithm for tamper detection in text documents.
\newblock In {\em 2010 2nd International Conference on Computer Engineering and Technology}, volume~6, pages V6--378. IEEE, 2010.

\bibitem{jalil2010content}
Zunera Jalil, Anwar~M Mirza, and Maria Sabir.
\newblock Content based zero-watermarking algorithm for authentication of text documents.
\newblock {\em arXiv preprint arXiv:1003.1796}, 2010.

\bibitem{ji2023beavertails}
Jiaming Ji, Mickel Liu, Juntao Dai, Xuehai Pan, Chi Zhang, Ce~Bian, Ruiyang Sun, Yizhou Wang, and Yaodong Yang.
\newblock Beavertails: Towards improved safety alignment of llm via a human-preference dataset.
\newblock {\em arXiv preprint arXiv:2307.04657}, 2023.

\bibitem{jiang2023low}
Jingang Jiang, Xiangyang Liu, and Chenyou Fan.
\newblock Low-parameter federated learning with large language models.
\newblock {\em arXiv preprint arXiv:2307.13896}, 2023.

\bibitem{jin2023binary}
Xin Jin, Jonathan Larson, Weiwei Yang, and Zhiqiang Lin.
\newblock Binary code summarization: Benchmarking chatgpt/gpt-4 and other large language models.
\newblock {\em arXiv preprint arXiv:2312.09601}, 2023.

\bibitem{jin2023better}
Yiqiao Jin, Mohit Chandra, Gaurav Verma, Yibo Hu, Munmun De~Choudhury, and Srijan Kumar.
\newblock Better to ask in english: Cross-lingual evaluation of large language models for healthcare queries.
\newblock {\em arXiv:2310.13132}, 2023.

\bibitem{jin2024mm}
Yiqiao Jin, Minje Choi, Gaurav Verma, Jindong Wang, and Srijan Kumar.
\newblock Mm-soc: Benchmarking multimodal large language models in social media platforms.
\newblock {\em arXiv:2402.14154}, 2024.

\bibitem{jin2022towards}
Yiqiao Jin, Xiting Wang, Ruichao Yang, Yizhou Sun, Wei Wang, Hao Liao, and Xing Xie.
\newblock Towards fine-grained reasoning for fake news detection.
\newblock In {\em Proceedings of the AAAI Conference on Artificial Intelligence}, volume~36, pages 5746--5754, 2022.

\bibitem{kairouz2021advances}
Peter Kairouz, H~Brendan McMahan, Brendan Avent, Aur{\'e}lien Bellet, Mehdi Bennis, Arjun~Nitin Bhagoji, Kallista Bonawitz, Zachary Charles, Graham Cormode, Rachel Cummings, et~al.
\newblock Advances and open problems in federated learning.
\newblock {\em Foundations and Trends{\textregistered} in Machine Learning}, 14(1--2):1--210, 2021.

\bibitem{kamiran2012data}
Faisal Kamiran and Toon Calders.
\newblock Data preprocessing techniques for classification without discrimination.
\newblock {\em Knowledge and information systems}, 33(1):1--33, 2012.

\bibitem{kandpal2023backdoor}
Nikhil Kandpal, Matthew Jagielski, Florian Tram{\`e}r, and Nicholas Carlini.
\newblock Backdoor attacks for in-context learning with language models.
\newblock {\em arXiv preprint arXiv:2307.14692}, 2023.

\bibitem{kandpal2022deduplicating}
Nikhil Kandpal, Eric Wallace, and Colin Raffel.
\newblock Deduplicating training data mitigates privacy risks in language models.
\newblock In {\em International Conference on Machine Learning}, pages 10697--10707. PMLR, 2022.

\bibitem{kaneko2022unmasking}
Masahiro Kaneko and Danushka Bollegala.
\newblock Unmasking the mask--evaluating social biases in masked language models.
\newblock In {\em Proceedings of the AAAI Conference on Artificial Intelligence}, volume~36, pages 11954--11962, 2022.

\bibitem{kang2023impact}
Cheongwoong Kang and Jaesik Choi.
\newblock Impact of co-occurrence on factual knowledge of large language models.
\newblock {\em arXiv preprint arXiv:2310.08256}, 2023.

\bibitem{kharitonov2021text}
Eugene Kharitonov, Ann Lee, Adam Polyak, Yossi Adi, Jade Copet, Kushal Lakhotia, Tu-Anh Nguyen, Morgane Rivi{\`e}re, Abdelrahman Mohamed, Emmanuel Dupoux, et~al.
\newblock Text-free prosody-aware generative spoken language modeling.
\newblock {\em arXiv preprint arXiv:2109.03264}, 2021.

\bibitem{kim2022prosocialdialog}
Hyunwoo Kim, Youngjae Yu, Liwei Jiang, Ximing Lu, Daniel Khashabi, Gunhee Kim, Yejin Choi, and Maarten Sap.
\newblock {P}rosocial{D}ialog: A prosocial backbone for conversational agents.
\newblock In {\em Proceedings of the 2022 Conference on Empirical Methods in Natural Language Processing}, pages 4005--4029, Abu Dhabi, United Arab Emirates, December 2022. Association for Computational Linguistics.

\bibitem{kim2019multiaccuracy}
Michael~P Kim, Amirata Ghorbani, and James Zou.
\newblock Multiaccuracy: Black-box post-processing for fairness in classification.
\newblock In {\em Proceedings of the 2019 AAAI/ACM Conference on AI, Ethics, and Society}, pages 247--254, 2019.

\bibitem{kirchenbauer2023watermark}
John Kirchenbauer, Jonas Geiping, Yuxin Wen, Jonathan Katz, Ian Miers, and Tom Goldstein.
\newblock A watermark for large language models.
\newblock {\em arXiv preprint arXiv:2301.10226}, 2023.

\bibitem{kourou2015machine}
Konstantina Kourou, Themis~P Exarchos, Konstantinos~P Exarchos, Michalis~V Karamouzis, and Dimitrios~I Fotiadis.
\newblock Machine learning applications in cancer prognosis and prediction.
\newblock {\em Computational and structural biotechnology journal}, 13:8--17, 2015.

\bibitem{kuang2023federatedscope}
Weirui Kuang, Bingchen Qian, Zitao Li, Daoyuan Chen, Dawei Gao, Xuchen Pan, Yuexiang Xie, Yaliang Li, Bolin Ding, and Jingren Zhou.
\newblock Federatedscope-llm: A comprehensive package for fine-tuning large language models in federated learning.
\newblock {\em arXiv preprint arXiv:2309.00363}, 2023.

\bibitem{Kuditipudi2023RobustDW}
Rohith Kuditipudi, John Thickstun, Tatsunori Hashimoto, and Percy Liang.
\newblock Robust distortion-free watermarks for language models.
\newblock {\em ArXiv}, abs/2307.15593, 2023.

\bibitem{kumar2023advances}
Srijan Kumar.
\newblock Advances in ai for web integrity, equity, and well-being.
\newblock {\em Frontiers in Big Data}, 6:1125083, 2023.

\bibitem{lai2023large}
Jinqi Lai, Wensheng Gan, Jiayang Wu, Zhenlian Qi, and Philip~S Yu.
\newblock Large language models in law: A survey.
\newblock {\em arXiv preprint arXiv:2312.03718}, 2023.

\bibitem{lee2021deduplicating}
Katherine Lee, Daphne Ippolito, Andrew Nystrom, Chiyuan Zhang, Douglas Eck, Chris Callison-Burch, and Nicholas Carlini.
\newblock Deduplicating training data makes language models better.
\newblock {\em arXiv preprint arXiv:2107.06499}, 2021.

\bibitem{lee2022factuality}
Nayeon Lee, Wei Ping, Peng Xu, Mostofa Patwary, Pascale~N Fung, Mohammad Shoeybi, and Bryan Catanzaro.
\newblock Factuality enhanced language models for open-ended text generation.
\newblock {\em Advances in Neural Information Processing Systems}, 35:34586--34599, 2022.

\bibitem{li2022mpcformer}
Dacheng Li, Rulin Shao, Hongyi Wang, Han Guo, Eric~P Xing, and Hao Zhang.
\newblock Mpcformer: fast, performant and private transformer inference with mpc.
\newblock {\em arXiv preprint arXiv:2211.01452}, 2022.

\bibitem{li2024uv}
Hong Li, Yutang Feng, Song Xue, Xuhui Liu, Bohan Zeng, Shanglin Li, Boyu Liu, Jianzhuang Liu, Shumin Han, and Baochang Zhang.
\newblock Uv-idm: Identity-conditioned latent diffusion model for face uv-texture generation.
\newblock In {\em CVPR}, pages 10585--10595, 2024.

\bibitem{li2023generative}
Junlong Li, Shichao Sun, Weizhe Yuan, Run-Ze Fan, Hai Zhao, and Pengfei Liu.
\newblock Generative judge for evaluating alignment.
\newblock {\em arXiv preprint arXiv:2310.05470}, 2023.

\bibitem{li2023halueval}
Junyi Li, Xiaoxue Cheng, Wayne~Xin Zhao, Jian-Yun Nie, and Ji-Rong Wen.
\newblock Halueval: A large-scale hallucination evaluation benchmark for large language models.
\newblock In {\em EMNLP}, pages 6449--6464, 2023.

\bibitem{li2023mope}
Marvin Li, Jason Wang, Jeffrey Wang, and Seth Neel.
\newblock Mope: Model perturbation-based privacy attacks on language models.
\newblock {\em arXiv preprint arXiv:2310.14369}, 2023.

\bibitem{li2021prefix}
Xiang~Lisa Li and Percy Liang.
\newblock Prefix-tuning: Optimizing continuous prompts for generation.
\newblock In {\em Proceedings of the 59th Annual Meeting of the Association for Computational Linguistics and the 11th International Joint Conference on Natural Language Processing (Volume 1: Long Papers)}, pages 4582--4597, Online, August 2021. Association for Computational Linguistics.

\bibitem{li2021large}
Xuechen Li, Florian Tramer, Percy Liang, and Tatsunori Hashimoto.
\newblock Large language models can be strong differentially private learners.
\newblock {\em arXiv preprint arXiv:2110.05679}, 2021.

\bibitem{li2023privacy}
Yansong Li, Zhixing Tan, and Yang Liu.
\newblock Privacy-preserving prompt tuning for large language model services.
\newblock {\em arXiv preprint arXiv:2305.06212}, 2023.

\bibitem{li2022backdoor}
Yiming Li, Yong Jiang, Zhifeng Li, and Shu-Tao Xia.
\newblock Backdoor learning: A survey.
\newblock {\em IEEE Transactions on Neural Networks and Learning Systems}, 2022.

\bibitem{li2023prompt}
Yingji Li, Mengnan Du, Xin Wang, and Ying Wang.
\newblock Prompt tuning pushes farther, contrastive learning pulls closer: A two-stage approach to mitigate social biases.
\newblock In {\em Proceedings of the 61st Annual Meeting of the Association for Computational Linguistics (Volume 1: Long Papers)}, pages 14254--14267, Toronto, Canada, July 2023. Association for Computational Linguistics.

\bibitem{li2023batgpt}
Zuchao Li, Shitou Zhang, Hai Zhao, Yifei Yang, and Dongjie Yang.
\newblock Batgpt: A bidirectional autoregessive talker from generative pre-trained transformer.
\newblock {\em arXiv preprint arXiv:2307.00360}, 2023.

\bibitem{lian2024public}
Ying Lian, Huiting Tang, Mengting Xiang, and Xuefan Dong.
\newblock Public attitudes and sentiments toward chatgpt in china: A text mining analysis based on social media.
\newblock {\em Technology in Society}, 76:102442, 2024.

\bibitem{liang2021towards}
Paul~Pu Liang, Chiyu Wu, Louis-Philippe Morency, and Ruslan Salakhutdinov.
\newblock Towards understanding and mitigating social biases in language models.
\newblock In {\em International Conference on Machine Learning}, pages 6565--6576. PMLR, 2021.

\bibitem{liang2022holistic}
Percy Liang, Rishi Bommasani, Tony Lee, Dimitris Tsipras, Dilara Soylu, Michihiro Yasunaga, Yian Zhang, Deepak Narayanan, Yuhuai Wu, Ananya Kumar, et~al.
\newblock Holistic evaluation of language models.
\newblock {\em arXiv preprint arXiv:2211.09110}, 2022.

\bibitem{liang2023merge}
Zi~Liang, Pinghui Wang, Ruofei Zhang, Nuo Xu, and Shuo Zhang.
\newblock Merge: Fast private text generation.
\newblock {\em arXiv preprint arXiv:2305.15769}, 2023.

\bibitem{Liu2023APW}
Aiwei Liu, Leyi Pan, Xuming Hu, Shuang Li, Lijie Wen, Irwin King, and Philip~S. Yu.
\newblock A private watermark for large language models.
\newblock {\em ArXiv}, abs/2307.16230, 2023.

\bibitem{liu2023hallusionbench}
Fuxiao Liu, Tianrui Guan, Zongxia Li, Lichang Chen, Yaser Yacoob, Dinesh Manocha, and Tianyi Zhou.
\newblock Hallusionbench: You see what you think? or you think what you see? an image-context reasoning benchmark challenging for gpt-4v (ision), llava-1.5, and other multi-modality models.
\newblock {\em arXiv:2310.14566}, 2023.

\bibitem{liu2023mitigating}
Fuxiao Liu, Kevin Lin, Linjie Li, Jianfeng Wang, Yaser Yacoob, and Lijuan Wang.
\newblock Mitigating hallucination in large multi-modal models via robust instruction tuning.
\newblock In {\em The Twelfth International Conference on Learning Representations}, 2023.

\bibitem{liu2023mmc}
Fuxiao Liu, Xiaoyang Wang, Wenlin Yao, Jianshu Chen, Kaiqiang Song, Sangwoo Cho, Yaser Yacoob, and Dong Yu.
\newblock Mmc: Advancing multimodal chart understanding with large-scale instruction tuning.
\newblock {\em arXiv preprint arXiv:2311.10774}, 2023.

\bibitem{liu2023covid}
Fuxiao Liu, Yaser Yacoob, and Abhinav Shrivastava.
\newblock Covid-vts: Fact extraction and verification on short video platforms.
\newblock {\em arXiv preprint arXiv:2302.07919}, 2023.

\bibitem{liu2022trainondata}
Gaoyang Liu, Tianlong Xu, Xiaoqiang Ma, and Chen Wang.
\newblock Your model trains on my data? protecting intellectual property of training data via membership fingerprint authentication.
\newblock {\em IEEE Transactions on Information Forensics and Security}, 17:1024--1037, 2022.

\bibitem{liu2019does}
Haochen Liu, Jamell Dacon, Wenqi Fan, Hui Liu, Zitao Liu, and Jiliang Tang.
\newblock Does gender matter? towards fairness in dialogue systems.
\newblock {\em arXiv preprint arXiv:1910.10486}, 2019.

\bibitem{liu2021mitigating}
Ruibo Liu, Chenyan Jia, Jason Wei, Guangxuan Xu, Lili Wang, and Soroush Vosoughi.
\newblock Mitigating political bias in language models through reinforced calibration.
\newblock In {\em Proceedings of the AAAI Conference on Artificial Intelligence}, volume~35, pages 14857--14866, 2021.

\bibitem{liu2024does}
Shengchao Liu, Xiaoming Liu, Yichen Wang, Zehua Cheng, Chengzhengxu Li, Zhaohan Zhang, Yu~Lan, and Chao Shen.
\newblock Does$\backslash$textsc $\{$DetectGPT$\}$ fully utilize perturbation? selective perturbation on model-based contrastive learning detector would be better.
\newblock {\em arXiv preprint arXiv:2402.00263}, 2024.

\bibitem{liu2023gpt}
Xiao Liu, Yanan Zheng, Zhengxiao Du, Ming Ding, Yujie Qian, Zhilin Yang, and Jie Tang.
\newblock Gpt understands, too.
\newblock {\em AI Open}, 2023.

\bibitem{liu2023coco}
Xiaoming Liu, Zhaohan Zhang, Yichen Wang, Hang Pu, Yu~Lan, and Chao Shen.
\newblock Coco: Coherence-enhanced machine-generated text detection under low resource with contrastive learning.
\newblock In {\em Proceedings of the 2023 Conference on Empirical Methods in Natural Language Processing}, pages 16167--16188, 2023.

\bibitem{liu2023trustworthy}
Yang Liu, Yuanshun Yao, Jean-Francois Ton, Xiaoying Zhang, Ruocheng Guo~Hao Cheng, Yegor Klochkov, Muhammad~Faaiz Taufiq, and Hang Li.
\newblock Trustworthy llms: a survey and guideline for evaluating large language models' alignment.
\newblock {\em arXiv preprint arXiv:2308.05374}, 2023.

\bibitem{liu2019roberta}
Yinhan Liu, Myle Ott, Naman Goyal, Jingfei Du, Mandar Joshi, Danqi Chen, Omer Levy, Mike Lewis, Luke Zettlemoyer, and Veselin Stoyanov.
\newblock Ro{BERT}a: A robustly optimized bert pretraining approach.
\newblock {\em arXiv preprint arXiv:1907.11692}, 2019.

\bibitem{liu2023watermarking}
Yixin Liu, Hongsheng Hu, Xuyun Zhang, and Lichao Sun.
\newblock Watermarking text data on large language models for dataset copyright protection.
\newblock {\em arXiv preprint arXiv:2305.13257}, 2023.

\bibitem{liu2022stolenencoder}
Yupei Liu, Jinyuan Jia, Hongbin Liu, and Neil~Zhenqiang Gong.
\newblock Stolenencoder: stealing pre-trained encoders in self-supervised learning.
\newblock In {\em Proceedings of the 2022 ACM SIGSAC Conference on Computer and Communications Security}, pages 2115--2128, 2022.

\bibitem{liu2023breaking}
Zheyuan Liu, Guangyao Dou, Yijun Tian, Chunhui Zhang, Eli Chien, and Ziwei Zhu.
\newblock Breaking the trilemma of privacy, utility, efficiency via controllable machine unlearning.
\newblock {\em arXiv preprint arXiv:2310.18574}, 2023.

\bibitem{lu2022quark}
Ximing Lu, Sean Welleck, Jack Hessel, Liwei Jiang, Lianhui Qin, Peter West, Prithviraj Ammanabrolu, and Yejin Choi.
\newblock Quark: Controllable text generation with reinforced unlearning.
\newblock {\em Advances in neural information processing systems}, 35:27591--27609, 2022.

\bibitem{lucas2023gpts}
Evan Lucas and Timothy Havens.
\newblock Gpts don’t keep secrets: Searching for backdoor watermark triggers in autoregressive language models.
\newblock In {\em Proceedings of the 3rd Workshop on Trustworthy Natural Language Processing (TrustNLP 2023)}, pages 242--248, 2023.

\bibitem{lyu2024task}
Weimin Lyu, Xiao Lin, Songzhu Zheng, Lu~Pang, Haibin Ling, Susmit Jha, and Chao Chen.
\newblock Task-agnostic detector for insertion-based backdoor attacks.
\newblock {\em arXiv preprint arXiv:2403.17155}, 2024.

\bibitem{lyu2023backdoor}
Weimin Lyu, Songzhu Zheng, Haibin Ling, and Chao Chen.
\newblock Backdoor attacks against transformers with attention enhancement.
\newblock In {\em ICLR 2023 Workshop on Backdoor Attacks and Defenses in Machine Learning}, 2023.

\bibitem{lyu2022study}
Weimin Lyu, Songzhu Zheng, Tengfei Ma, and Chao Chen.
\newblock A study of the attention abnormality in trojaned berts.
\newblock In {\em Proceedings of the 2022 Conference of the North American Chapter of the Association for Computational Linguistics: Human Language Technologies}, pages 4727--4741, 2022.

\bibitem{lyu2023attention}
Weimin Lyu, Songzhu Zheng, Lu~Pang, Haibin Ling, and Chao Chen.
\newblock Attention-enhancing backdoor attacks against bert-based models.
\newblock In {\em Findings of the Association for Computational Linguistics: EMNLP 2023}, pages 10672--10690, 2023.

\bibitem{mai2023split}
Peihua Mai, Ran Yan, Zhe Huang, Youjia Yang, and Yan Pang.
\newblock Split-and-denoise: Protect large language model inference with local differential privacy.
\newblock {\em arXiv preprint arXiv:2310.09130}, 2023.

\bibitem{maini2021dataset}
Pratyush Maini, Mohammad Yaghini, and Nicolas Papernot.
\newblock Dataset inference: Ownership resolution in machine learning.
\newblock {\em arXiv preprint arXiv:2104.10706}, 2021.

\bibitem{majmudar2022differentially}
Jimit Majmudar, Christophe Dupuy, Charith Peris, Sami Smaili, Rahul Gupta, and Richard Zemel.
\newblock Differentially private decoding in large language models.
\newblock {\em arXiv preprint arXiv:2205.13621}, 2022.

\bibitem{mao2024compressibility}
Yu~Mao, Weilan Wang, Hongchao Du, Nan Guan, and Chun~Jason Xue.
\newblock On the compressibility of quantized large language models.
\newblock {\em arXiv preprint arXiv:2403.01384}, 2024.

\bibitem{mattern2023membership}
Justus Mattern, Fatemehsadat Mireshghallah, Zhijing Jin, Bernhard Sch{\"o}lkopf, Mrinmaya Sachan, and Taylor Berg-Kirkpatrick.
\newblock Membership inference attacks against language models via neighbourhood comparison.
\newblock {\em arXiv preprint arXiv:2305.18462}, 2023.

\bibitem{maudslay2019s}
Rowan~Hall Maudslay, Hila Gonen, Ryan Cotterell, and Simone Teufel.
\newblock It’s all in the name: Mitigating gender bias with name-based counterfactual data substitution.
\newblock In {\em Proceedings of the 2019 Conference on Empirical Methods in Natural Language Processing and the 9th International Joint Conference on Natural Language Processing (EMNLP-IJCNLP)}, pages 5267--5275, 2019.

\bibitem{may2019measuring}
Chandler May, Alex Wang, Shikha Bordia, Samuel~R. Bowman, and Rachel Rudinger.
\newblock On measuring social biases in sentence encoders.
\newblock In {\em Proceedings of the 2019 Conference of the North {A}merican Chapter of the Association for Computational Linguistics: Human Language Technologies, Volume 1 (Long and Short Papers)}, pages 622--628, Minneapolis, Minnesota, June 2019. Association for Computational Linguistics.

\bibitem{mccoy2023much}
R~Thomas McCoy, Paul Smolensky, Tal Linzen, Jianfeng Gao, and Asli Celikyilmaz.
\newblock How much do language models copy from their training data? evaluating linguistic novelty in text generation using raven.
\newblock {\em Transactions of the Association for Computational Linguistics}, 11:652--670, 2023.

\bibitem{mcmahan2017learning}
H~Brendan McMahan, Daniel Ramage, Kunal Talwar, and Li~Zhang.
\newblock Learning differentially private recurrent language models.
\newblock {\em arXiv preprint arXiv:1710.06963}, 2017.

\bibitem{meade2023using}
Nicholas Meade, Spandana Gella, Devamanyu Hazarika, Prakhar Gupta, Di~Jin, Siva Reddy, Yang Liu, and Dilek Hakkani-T{\"u}r.
\newblock Using in-context learning to improve dialogue safety.
\newblock {\em arXiv preprint arXiv:2302.00871}, 2023.

\bibitem{mechura2022taxonomy}
Michal M{\v{e}}chura.
\newblock A taxonomy of bias-causing ambiguities in machine translation.
\newblock In {\em Proceedings of the 4th Workshop on Gender Bias in Natural Language Processing (GeBNLP)}, pages 168--173, Seattle, Washington, July 2022. Association for Computational Linguistics.

\bibitem{meeus2023concerns}
Matthieu Meeus, Shubham Jain, and Yves-Alexandre de~Montjoye.
\newblock Concerns about using a digital mask to safeguard patient privacy.
\newblock {\em Nature Medicine}, 29(7):1658--1659, 2023.

\bibitem{mehrabi2021survey}
Ninareh Mehrabi, Fred Morstatter, Nripsuta Saxena, Kristina Lerman, and Aram Galstyan.
\newblock A survey on bias and fairness in machine learning.
\newblock {\em ACM Computing Surveys (CSUR)}, 54(6):1--35, 2021.

\bibitem{melis2019exploiting}
Luca Melis, Congzheng Song, Emiliano De~Cristofaro, and Vitaly Shmatikov.
\newblock Exploiting unintended feature leakage in collaborative learning.
\newblock In {\em 2019 IEEE symposium on security and privacy (SP)}, pages 691--706. IEEE, 2019.

\bibitem{mesko2023imperative}
Bertalan Mesk{\'o} and Eric~J Topol.
\newblock The imperative for regulatory oversight of large language models (or generative ai) in healthcare.
\newblock {\em npj Digital Medicine}, 6(1):120, 2023.

\bibitem{minssen2023challenges}
Timo Minssen, Effy Vayena, and I~Glenn Cohen.
\newblock The challenges for regulating medical use of chatgpt and other large language models.
\newblock {\em Jama}, 2023.

\bibitem{Mir2014CopyrightFW}
Nighat Mir.
\newblock Copyright for web content using invisible text watermarking.
\newblock {\em Comput. Hum. Behav.}, 30:648--653, 2014.

\bibitem{mireshghallah2022quantifying}
Fatemehsadat Mireshghallah, Kartik Goyal, Archit Uniyal, Taylor Berg-Kirkpatrick, and Reza Shokri.
\newblock Quantifying privacy risks of masked language models using membership inference attacks.
\newblock {\em arXiv preprint arXiv:2203.03929}, 2022.

\bibitem{mitchell2023detectgpt}
Eric Mitchell, Yoonho Lee, Alexander Khazatsky, Christopher~D Manning, and Chelsea Finn.
\newblock Detectgpt: Zero-shot machine-generated text detection using probability curvature.
\newblock In {\em International Conference on Machine Learning}, pages 24950--24962. PMLR, 2023.

\bibitem{mitchell2022memory}
Eric Mitchell, Charles Lin, Antoine Bosselut, Christopher~D Manning, and Chelsea Finn.
\newblock Memory-based model editing at scale.
\newblock In {\em International Conference on Machine Learning}, pages 15817--15831. PMLR, 2022.

\bibitem{mo2024large}
Yuhong Mo, Hao Qin, Yushan Dong, Ziyi Zhu, and Zhenglin Li.
\newblock Large language model (llm) ai text generation detection based on transformer deep learning algorithm.
\newblock {\em arXiv preprint arXiv:2405.06652}, 2024.

\bibitem{mozafari2020hate}
Marzieh Mozafari, Reza Farahbakhsh, and No{\"e}l Crespi.
\newblock Hate speech detection and racial bias mitigation in social media based on bert model.
\newblock {\em PloS one}, 15(8):e0237861, 2020.

\bibitem{mugunthan2019smpai}
Vaikkunth Mugunthan, Antigoni Polychroniadou, David Byrd, and Tucker~Hybinette Balch.
\newblock Smpai: Secure multi-party computation for federated learning.
\newblock In {\em Proceedings of the NeurIPS 2019 Workshop on Robust AI in Financial Services}, pages 1--9. MIT Press Cambridge, MA, USA, 2019.

\bibitem{nangia2020crows}
Nikita Nangia, Clara Vania, Rasika Bhalerao, and Samuel~R. Bowman.
\newblock {CrowS-Pairs: A Challenge Dataset for Measuring Social Biases in Masked Language Models}.
\newblock In {\em Proceedings of the 2020 Conference on Empirical Methods in Natural Language Processing}, pages 1953--1967, Online, November 2020. Association for Computational Linguistics.

\bibitem{venkit2023nationality}
Pranav Narayanan~Venkit, Sanjana Gautam, Ruchi Panchanadikar, Ting-Hao Huang, and Shomir Wilson.
\newblock Nationality bias in text generation.
\newblock In {\em Proceedings of the 17th Conference of the European Chapter of the Association for Computational Linguistics}, pages 116--122, Dubrovnik, Croatia, May 2023. Association for Computational Linguistics.

\bibitem{nasr2023scalable}
Milad Nasr, Nicholas Carlini, Jonathan Hayase, Matthew Jagielski, A~Feder Cooper, Daphne Ippolito, Christopher~A Choquette-Choo, Eric Wallace, Florian Tram{\`e}r, and Katherine Lee.
\newblock Scalable extraction of training data from (production) language models.
\newblock {\em arXiv preprint arXiv:2311.17035}, 2023.

\bibitem{neo2024towards}
Neng Kai~Nigel Neo, Yeon-Chang Lee, Yiqiao Jin, Sang-Wook Kim, and Srijan Kumar.
\newblock Towards fair graph anomaly detection: Problem, new datasets, and evaluation.
\newblock {\em arXiv:2402.15988}, 2024.

\bibitem{nguyen2020input}
Tuan~Anh Nguyen and Anh Tran.
\newblock Input-aware dynamic backdoor attack.
\newblock In {\em Advances in Neural Information Processing Systems}, 2020.

\bibitem{nori2023capabilities}
Harsha Nori, Nicholas King, Scott~Mayer McKinney, Dean Carignan, and Eric Horvitz.
\newblock Capabilities of gpt-4 on medical challenge problems.
\newblock {\em arXiv:2303.13375}, 2023.

\bibitem{nozza2021honest}
Debora Nozza, Federico Bianchi, and Dirk Hovy.
\newblock {HONEST}: Measuring hurtful sentence completion in language models.
\newblock In {\em Proceedings of the 2021 Conference of the North American Chapter of the Association for Computational Linguistics: Human Language Technologies}, pages 2398--2406, Online, June 2021. Association for Computational Linguistics, Association for Computational Linguistics.

\bibitem{oh2022learning}
Changdae Oh, Heeji Won, Junhyuk So, Taero Kim, Yewon Kim, Hosik Choi, and Kyungwoo Song.
\newblock Learning fair representation via distributional contrastive disentanglement.
\newblock In {\em Proceedings of the 28th ACM SIGKDD Conference on Knowledge Discovery and Data Mining}, pages 1295--1305, 2022.

\bibitem{orgad2023blind}
Hadas Orgad and Yonatan Belinkov.
\newblock {BLIND}: Bias removal with no demographics.
\newblock In {\em Proceedings of the 61st Annual Meeting of the Association for Computational Linguistics (Volume 1: Long Papers)}, pages 8801--8821, Toronto, Canada, July 2023. Association for Computational Linguistics.

\bibitem{ouyang2022training}
Long Ouyang, Jeffrey Wu, Xu~Jiang, Diogo Almeida, Carroll Wainwright, Pamela Mishkin, Chong Zhang, Sandhini Agarwal, Katarina Slama, Alex Ray, et~al.
\newblock Training language models to follow instructions with human feedback.
\newblock {\em Advances in Neural Information Processing Systems}, 35:27730--27744, 2022.

\bibitem{pan2023automatically}
Liangming Pan, Michael Saxon, Wenda Xu, Deepak Nathani, Xinyi Wang, and William~Yang Wang.
\newblock Automatically correcting large language models: Surveying the landscape of diverse self-correction strategies.
\newblock {\em arXiv preprint arXiv:2308.03188}, 2023.

\bibitem{park2023never}
SunYoung Park, Kyuri Choi, Haeun Yu, and Youngjoong Ko.
\newblock Never too late to learn: Regularizing gender bias in coreference resolution.
\newblock In {\em Proceedings of the Sixteenth ACM International Conference on Web Search and Data Mining}, WSDM '23, page 15–23, New York, NY, USA, 2023. Association for Computing Machinery.

\bibitem{parrish2022bbq}
Alicia Parrish, Angelica Chen, Nikita Nangia, Vishakh Padmakumar, Jason Phang, Jana Thompson, Phu~Mon Htut, and Samuel Bowman.
\newblock {BBQ}: A hand-built bias benchmark for question answering.
\newblock In {\em Findings of the Association for Computational Linguistics: ACL 2022}, pages 2086--2105, Dublin, Ireland, May 2022. Association for Computational Linguistics.

\bibitem{patrocinio2023artificial}
J{\'u}lio C{\'e}sar~Parente Patroc{\'\i}nio and D{\'e}bora Barreto~Santana de~Andrade.
\newblock Artificial intelligence, algorithmic recommendation and decision-making in european union law:: analysis of the regulatory challenge and legal certainty.
\newblock {\em Latin American Center of European Studies}, 3(2):136--179, 2023.

\bibitem{pawelczyk2023context}
Martin Pawelczyk, Seth Neel, and Himabindu Lakkaraju.
\newblock In-context unlearning: Language models as few shot unlearners.
\newblock {\em arXiv preprint arXiv:2310.07579}, 2023.

\bibitem{peng2023check}
Baolin Peng, Michel Galley, Pengcheng He, Hao Cheng, Yujia Xie, Yu~Hu, Qiuyuan Huang, Lars Liden, Zhou Yu, Weizhu Chen, et~al.
\newblock Check your facts and try again: Improving large language models with external knowledge and automated feedback.
\newblock {\em arXiv preprint arXiv:2302.12813}, 2023.

\bibitem{peng2023you}
Wenjun Peng, Jingwei Yi, Fangzhao Wu, Shangxi Wu, Bin Zhu, Lingjuan Lyu, Binxing Jiao, Tong Xu, Guangzhong Sun, and Xing Xie.
\newblock Are you copying my model? protecting the copyright of large language models for eaas via backdoor watermark.
\newblock {\em arXiv preprint arXiv:2305.10036}, 2023.

\bibitem{pessach2024fairness}
Dana Pessach, Tamir Tassa, and Erez Shmueli.
\newblock Fairness-driven private collaborative machine learning.
\newblock {\em ACM Transactions on Intelligent Systems and Technology}, 15(2):1--30, 2024.

\bibitem{petersen2021post}
Felix Petersen, Debarghya Mukherjee, Yuekai Sun, and Mikhail Yurochkin.
\newblock Post-processing for individual fairness.
\newblock {\em Advances in Neural Information Processing Systems}, 34:25944--25955, 2021.

\bibitem{petitcolas1999infohide}
F.A.P. Petitcolas, R.J. Anderson, and M.G. Kuhn.
\newblock Information hiding-a survey.
\newblock {\em Proceedings of the IEEE}, 87(7):1062--1078, 1999.

\bibitem{qian2022perturbation}
Rebecca Qian, Candace Ross, Jude Fernandes, Eric~Michael Smith, Douwe Kiela, and Adina Williams.
\newblock Perturbation augmentation for fairer {NLP}.
\newblock In {\em Proceedings of the 2022 Conference on Empirical Methods in Natural Language Processing}, pages 9496--9521, Abu Dhabi, United Arab Emirates, December 2022. Association for Computational Linguistics.

\bibitem{qian2019reducing}
Yusu Qian, Urwa Muaz, Ben Zhang, and Jae~Won Hyun.
\newblock Reducing gender bias in word-level language models with a gender-equalizing loss function.
\newblock In {\em Proceedings of the 57th Annual Meeting of the Association for Computational Linguistics: Student Research Workshop}, pages 223--228, Florence, Italy, July 2019. Association for Computational Linguistics.

\bibitem{radford2018improving}
Alec Radford, Karthik Narasimhan, Tim Salimans, Ilya Sutskever, et~al.
\newblock Improving language understanding by generative pre-training, 2018.

\bibitem{radford2019language}
Alec Radford, Jeffrey Wu, Rewon Child, David Luan, Dario Amodei, Ilya Sutskever, et~al.
\newblock Language models are unsupervised multitask learners.
\newblock {\em OpenAI blog}, 1(8):9, 2019.

\bibitem{raji2022outsider}
Inioluwa~Deborah Raji, Peggy Xu, Colleen Honigsberg, and Daniel Ho.
\newblock Outsider oversight: Designing a third party audit ecosystem for ai governance.
\newblock In {\em Proceedings of the 2022 AAAI/ACM Conference on AI, Ethics, and Society}, pages 557--571, 2022.

\bibitem{rajpurkar2016squad}
Pranav Rajpurkar, Jian Zhang, Konstantin Lopyrev, and Percy Liang.
\newblock {SQ}u{AD}: 100,000+ questions for machine comprehension of text.
\newblock In {\em Proceedings of the 2016 Conference on Empirical Methods in Natural Language Processing}, pages 2383--2392, Austin, Texas, November 2016. Association for Computational Linguistics.

\bibitem{ram2023context}
Ori Ram, Yoav Levine, Itay Dalmedigos, Dor Muhlgay, Amnon Shashua, Kevin Leyton-Brown, and Yoav Shoham.
\newblock In-context retrieval-augmented language models.
\newblock {\em Transactions of the Association for Computational Linguistics}, 11:1316--1331, 2023.

\bibitem{rashid2023fltrojan}
Md~Rafi~Ur Rashid, Vishnu~Asutosh Dasu, Kang Gu, Najrin Sultana, and Shagufta Mehnaz.
\newblock Fltrojan: Privacy leakage attacks against federated language models through selective weight tampering.
\newblock {\em arXiv preprint arXiv:2310.16152}, 2023.

\bibitem{ravfogel2020null}
Shauli Ravfogel, Yanai Elazar, Hila Gonen, Michael Twiton, and Yoav Goldberg.
\newblock Null it out: Guarding protected attributes by iterative nullspace projection.
\newblock In {\em Proceedings of the 58th Annual Meeting of the Association for Computational Linguistics}, pages 7237--7256, Online, July 2020. Association for Computational Linguistics.

\bibitem{regulation2018general}
General Data~Protection Regulation.
\newblock General data protection regulation (gdpr).
\newblock {\em Intersoft Consulting, Accessed in October}, 24(1), 2018.

\bibitem{rekabsaz2021societal}
Navid Rekabsaz, Simone Kopeinik, and Markus Schedl.
\newblock Societal biases in retrieved contents: Measurement framework and adversarial mitigation of bert rankers.
\newblock In {\em Proceedings of the 44th International ACM SIGIR Conference on Research and Development in Information Retrieval}, pages 306--316, 2021.

\bibitem{rekabsaz2020do}
Navid Rekabsaz and Markus Schedl.
\newblock Do neural ranking models intensify gender bias?
\newblock In {\em Proceedings of the 43rd International ACM SIGIR Conference on Research and Development in Information Retrieval}, SIGIR '20, page 2065–2068, New York, NY, USA, 2020. Association for Computing Machinery.

\bibitem{ren2023robots}
Allen~Z Ren, Anushri Dixit, Alexandra Bodrova, Sumeet Singh, Stephen Tu, Noah Brown, Peng Xu, Leila Takayama, Fei Xia, Jake Varley, et~al.
\newblock Robots that ask for help: Uncertainty alignment for large language model planners.
\newblock {\em arXiv preprint arXiv:2307.01928}, 2023.

\bibitem{Ren2023ARS}
Jie Ren, Han Xu, Yiding Liu, Yingqian Cui, Shuaiqiang Wang, Dawei Yin, and Jiliang Tang.
\newblock A robust semantics-based watermark for large language model against paraphrasing.
\newblock {\em ArXiv}, abs/2311.08721, 2023.

\bibitem{roberts2021chinese}
Huw Roberts, Josh Cowls, Jessica Morley, Mariarosaria Taddeo, Vincent Wang, and Luciano Floridi.
\newblock The chinese approach to artificial intelligence: an analysis of policy, ethics, and regulation.
\newblock {\em AI \& society}, 36:59--77, 2021.

\bibitem{roberts2023breaking}
Jessica Roberts, Rachel Lowy, Huaigu Li, Jon Bellona, Leslie Smith, and Amy Bower.
\newblock Breaking down the visual barrier: Designing data interactions for the visually impaired in informal learning settings.
\newblock In {\em CSCL}. International Society of the Learning Sciences, 2023.

\bibitem{ruan2024s2e}
Kangrui Ruan, Xin He, Jiyang Wang, Xiaozhou Zhou, Helian Feng, and Ali Kebarighotbi.
\newblock S2e: Towards an end-to-end entity resolution solution from acoustic signal.
\newblock In {\em ICASSP 2024-2024 IEEE International Conference on Acoustics, Speech and Signal Processing (ICASSP)}, pages 10441--10445. IEEE, 2024.

\bibitem{saunders2022first}
Danielle Saunders, Rosie Sallis, and Bill Byrne.
\newblock First the worst: Finding better gender translations during beam search.
\newblock In {\em Findings of the Association for Computational Linguistics: ACL 2022}, pages 3814--3823, Dublin, Ireland, May 2022. Association for Computational Linguistics.

\bibitem{shi2021selective}
Weiyan Shi, Aiqi Cui, Evan Li, Ruoxi Jia, and Zhou Yu.
\newblock Selective differential privacy for language modeling.
\newblock {\em arXiv preprint arXiv:2108.12944}, 2021.

\bibitem{shoaib2023deepfakes}
Mohamed~R Shoaib, Zefan Wang, Milad~Taleby Ahvanooey, and Jun Zhao.
\newblock Deepfakes, misinformation, and disinformation in the era of frontier ai, generative ai, and large ai models.
\newblock In {\em ICCA}, pages 1--7. IEEE, 2023.

\bibitem{shokri2017membership}
Reza Shokri, Marco Stronati, Congzheng Song, and Vitaly Shmatikov.
\newblock Membership inference attacks against machine learning models.
\newblock In {\em 2017 IEEE symposium on security and privacy (SP)}, pages 3--18. IEEE, 2017.

\bibitem{Singh2013ASO}
Prabhishek Singh and Ramneet~Singh Chadha.
\newblock A survey of digital watermarking techniques, applications and attacks.
\newblock 2013.

\bibitem{smith2022im}
Eric~Michael Smith, Melissa Hall, Melanie Kambadur, Eleonora Presani, and Adina Williams.
\newblock {``}{I}{'}m sorry to hear that{''}: Finding new biases in language models with a holistic descriptor dataset.
\newblock In {\em Proceedings of the 2022 Conference on Empirical Methods in Natural Language Processing}, pages 9180--9211, Abu Dhabi, United Arab Emirates, December 2022. Association for Computational Linguistics.

\bibitem{smith2022m}
Eric~Michael Smith, Melissa Hall, Melanie Kambadur, Eleonora Presani, and Adina Williams.
\newblock “i’m sorry to hear that”: Finding new biases in language models with a holistic descriptor dataset.
\newblock In {\em Proceedings of the 2022 Conference on Empirical Methods in Natural Language Processing}, pages 9180--9211, 2022.

\bibitem{song2020information}
Congzheng Song and Ananth Raghunathan.
\newblock Information leakage in embedding models.
\newblock In {\em Proceedings of the 2020 ACM SIGSAC conference on computer and communications security}, pages 377--390, 2020.

\bibitem{song2023preference}
Feifan Song, Bowen Yu, Minghao Li, Haiyang Yu, Fei Huang, Yongbin Li, and Houfeng Wang.
\newblock Preference ranking optimization for human alignment.
\newblock {\em arXiv preprint arXiv:2306.17492}, 2023.

\bibitem{staab2023beyond}
Robin Staab, Mark Vero, Mislav Balunovi{\'c}, and Martin Vechev.
\newblock Beyond memorization: Violating privacy via inference with large language models.
\newblock {\em arXiv preprint arXiv:2310.07298}, 2023.

\bibitem{su2024large}
Jing Su, Chufeng Jiang, Xin Jin, Yuxin Qiao, Tingsong Xiao, Hongda Ma, Rong Wei, Zhi Jing, Jiajun Xu, and Junhong Lin.
\newblock Large language models for forecasting and anomaly detection: A systematic literature review.
\newblock {\em arXiv preprint arXiv:2402.10350}, 2024.

\bibitem{sun2023moraldial}
Hao Sun, Zhexin Zhang, Fei Mi, Yasheng Wang, Wei Liu, Jianwei Cui, Bin Wang, Qun Liu, and Minlie Huang.
\newblock {M}oral{D}ial: A framework to train and evaluate moral dialogue systems via moral discussions.
\newblock In {\em Proceedings of the 61st Annual Meeting of the Association for Computational Linguistics (Volume 1: Long Papers)}, pages 2213--2230, Toronto, Canada, July 2023. Association for Computational Linguistics.

\bibitem{sun2024trustllm}
Lichao Sun, Yue Huang, Haoran Wang, Siyuan Wu, Qihui Zhang, Chujie Gao, Yixin Huang, Wenhan Lyu, Yixuan Zhang, Xiner Li, et~al.
\newblock Trustllm: Trustworthiness in large language models.
\newblock {\em arXiv preprint arXiv:2401.05561}, 2024.

\bibitem{sun2021they}
Tony Sun, Kellie Webster, Apu Shah, William~Yang Wang, and Melvin Johnson.
\newblock They, them, theirs: Rewriting with gender-neutral english.
\newblock {\em arXiv preprint arXiv:2102.06788}, 2021.

\bibitem{sun2023principle}
Zhiqing Sun, Yikang Shen, Qinhong Zhou, Hongxin Zhang, Zhenfang Chen, David Cox, Yiming Yang, and Chuang Gan.
\newblock Principle-driven self-alignment of language models from scratch with minimal human supervision.
\newblock {\em arXiv preprint arXiv:2305.03047}, 2023.

\bibitem{suresh2021framework}
Harini Suresh and John Guttag.
\newblock A framework for understanding sources of harm throughout the machine learning life cycle.
\newblock {\em Equity and access in algorithms, mechanisms, and optimization}, pages 1--9, 2021.

\bibitem{tang2023privacy}
Xinyu Tang, Richard Shin, Huseyin~A Inan, Andre Manoel, Fatemehsadat Mireshghallah, Zinan Lin, Sivakanth Gopi, Janardhan Kulkarni, and Robert Sim.
\newblock Privacy-preserving in-context learning with differentially private few-shot generation.
\newblock {\em arXiv preprint arXiv:2309.11765}, 2023.

\bibitem{taori2023stanford}
Rohan Taori, Ishaan Gulrajani, Tianyi Zhang, Yann Dubois, Xuechen Li, Carlos Guestrin, Percy Liang, and Tatsunori~B Hashimoto.
\newblock Stanford alpaca: An instruction-following llama model, 2023.

\bibitem{tian2024gnp}
Yijun Tian, Huan Song, Zichen Wang, Haozhu Wang, Ziqing Hu, Fang Wang, Nitesh~V Chawla, and Panpan Xu.
\newblock Graph neural prompting with large language models.
\newblock In {\em AAAI}, 2024.

\bibitem{tirumala2022memorization}
Kushal Tirumala, Aram Markosyan, Luke Zettlemoyer, and Armen Aghajanyan.
\newblock Memorization without overfitting: Analyzing the training dynamics of large language models.
\newblock {\em Advances in Neural Information Processing Systems}, 35:38274--38290, 2022.

\bibitem{tokpo2022text}
Ewoenam~Kwaku Tokpo and Toon Calders.
\newblock Text style transfer for bias mitigation using masked language modeling.
\newblock In {\em Proceedings of the 2022 Conference of the North American Chapter of the Association for Computational Linguistics: Human Language Technologies: Student Research Workshop}, pages 163--171, Hybrid: Seattle, Washington + Online, July 2022. Association for Computational Linguistics.

\bibitem{tong2023privinfer}
Meng Tong, Kejiang Chen, Yuang Qi, Jie Zhang, Weiming Zhang, and Nenghai Yu.
\newblock Privinfer: Privacy-preserving inference for black-box large language model.
\newblock {\em arXiv preprint arXiv:2310.12214}, 2023.

\bibitem{Topkara2005NaturalLW}
Mercan Topkara, C{\"u}neyt~M. Taskiran, and Edward~J. Delp.
\newblock Natural language watermarking.
\newblock In {\em IS\&T/SPIE Electronic Imaging}, 2005.

\bibitem{tople2019analyzing}
Shruti Tople, Marc Brockschmidt, Boris K{\"o}pf, Olga Ohrimenko, and Santiago Zanella-B{\'e}guelin.
\newblock Analyzing privacy loss in updates of natural language models.
\newblock 2019.

\bibitem{touvron2023llama}
Hugo Touvron, Thibaut Lavril, Gautier Izacard, Xavier Martinet, Marie-Anne Lachaux, Timoth{\'e}e Lacroix, Baptiste Rozi{\`e}re, Naman Goyal, Eric Hambro, Faisal Azhar, et~al.
\newblock Llama: Open and efficient foundation language models.
\newblock {\em arXiv preprint arXiv:2302.13971}, 2023.

\bibitem{touvron2023llama2}
Hugo Touvron, Louis Martin, Kevin Stone, Peter Albert, Amjad Almahairi, Yasmine Babaei, Nikolay Bashlykov, Soumya Batra, Prajjwal Bhargava, Shruti Bhosale, et~al.
\newblock Llama 2: Open foundation and fine-tuned chat models, 2023.
\newblock {\em URL https://arxiv. org/abs/2307.09288}, 2023.

\bibitem{triguero2024general}
Isaac Triguero, Daniel Molina, Javier Poyatos, Javier Del~Ser, and Francisco Herrera.
\newblock General purpose artificial intelligence systems (gpais): Properties, definition, taxonomy, societal implications and responsible governance.
\newblock {\em Information Fusion}, 103:102135, 2024.

\bibitem{tripto2023ship}
Nafis~Irtiza Tripto, Saranya Venkatraman, Dominik Macko, Robert Moro, Ivan Srba, Adaku Uchendu, Thai Le, and Dongwon Lee.
\newblock A ship of theseus: Curious cases of paraphrasing in llm-generated texts.
\newblock {\em arXiv preprint arXiv:2311.08374}, 2023.

\bibitem{truex2019hybrid}
Stacey Truex, Nathalie Baracaldo, Ali Anwar, Thomas Steinke, Heiko Ludwig, Rui Zhang, and Yi~Zhou.
\newblock A hybrid approach to privacy-preserving federated learning.
\newblock In {\em Proceedings of the 12th ACM workshop on artificial intelligence and security}, pages 1--11, 2019.

\bibitem{vanmassenhove2021neutral}
Eva Vanmassenhove, Chris Emmery, and Dimitar Shterionov.
\newblock {N}eu{T}ral {R}ewriter: {A} rule-based and neural approach to automatic rewriting into gender neutral alternatives.
\newblock In {\em Proceedings of the 2021 Conference on Empirical Methods in Natural Language Processing}, pages 8940--8948, Online and Punta Cana, Dominican Republic, November 2021. Association for Computational Linguistics.

\bibitem{verma2024mysterious}
Gaurav Verma, Minje Choi, Kartik Sharma, Jamelle Watson-Daniels, Sejoon Oh, and Srijan Kumar.
\newblock Mysterious projections: Multimodal llms gain domain-specific visual capabilities without richer cross-modal projections.
\newblock {\em arXiv:2402.16832}, 2024.

\bibitem{voigt2017eu}
Paul Voigt and Axel Von~dem Bussche.
\newblock The eu general data protection regulation (gdpr).
\newblock {\em A Practical Guide, 1st Ed., Cham: Springer International Publishing}, 10(3152676):10--5555, 2017.

\bibitem{wan2023faithfulness}
David Wan, Mengwen Liu, Kathleen McKeown, Markus Dreyer, and Mohit Bansal.
\newblock Faithfulness-aware decoding strategies for abstractive summarization.
\newblock {\em arXiv preprint arXiv:2303.03278}, 2023.

\bibitem{wang2018glue}
Alex Wang, Amanpreet Singh, Julian Michael, Felix Hill, Omer Levy, and Samuel~R Bowman.
\newblock Glue: A multi-task benchmark and analysis platform for natural language understanding.
\newblock {\em arXiv preprint arXiv:1804.07461}, 2018.

\bibitem{wang2023decodingtrust}
Boxin Wang, Weixin Chen, Hengzhi Pei, Chulin Xie, Mintong Kang, Chenhui Zhang, Chejian Xu, Zidi Xiong, Ritik Dutta, Rylan Schaeffer, et~al.
\newblock Decodingtrust: A comprehensive assessment of trustworthiness in gpt models.
\newblock {\em arXiv preprint arXiv:2306.11698}, 2023.

\bibitem{wang2021adversarial}
Boxin Wang, Chejian Xu, Shuohang Wang, Zhe Gan, Yu~Cheng, Jianfeng Gao, Ahmed~Hassan Awadallah, and Bo~Li.
\newblock Adversarial glue: A multi-task benchmark for robustness evaluation of language models.
\newblock {\em arXiv preprint arXiv:2111.02840}, 2021.

\bibitem{wang2022global}
Haoxiang Wang, Yite Wang, Ruoyu Sun, and Bo~Li.
\newblock Global convergence of maml and theory-inspired neural architecture search for few-shot learning.
\newblock In {\em Proceedings of the IEEE/CVF conference on computer vision and pattern recognition}, pages 9797--9808, 2022.

\bibitem{wang2023robustness}
Jindong Wang, Xixu Hu, Wenxin Hou, Hao Chen, Runkai Zheng, Yidong Wang, Linyi Yang, Haojun Huang, Wei Ye, Xiubo Geng, et~al.
\newblock On the robustness of chatgpt: An adversarial and out-of-distribution perspective.
\newblock {\em arXiv preprint arXiv:2302.12095}, 2023.

\bibitem{wang2023evaluation}
Junyang Wang, Yiyang Zhou, Guohai Xu, Pengcheng Shi, Chenlin Zhao, Haiyang Xu, Qinghao Ye, Ming Yan, Ji~Zhang, Jihua Zhu, et~al.
\newblock Evaluation and analysis of hallucination in large vision-language models.
\newblock {\em arXiv:2308.15126}, 2023.

\bibitem{Wang2023TowardsCT}
Lean Wang, Wenkai Yang, Deli Chen, Haozhe Zhou, Yankai Lin, Fandong Meng, Jie Zhou, and Xu~Sun.
\newblock Towards codable text watermarking for large language models.
\newblock {\em ArXiv}, abs/2307.15992, 2023.

\bibitem{wang2024mementos}
Xiyao Wang, Yuhang Zhou, Xiaoyu Liu, Hongjin Lu, Yuancheng Xu, Feihong He, Jaehong Yoon, Taixi Lu, Gedas Bertasius, Mohit Bansal, et~al.
\newblock Mementos: A comprehensive benchmark for multimodal large language model reasoning over image sequences.
\newblock {\em arXiv preprint arXiv:2401.10529}, 2024.

\bibitem{wang2024stumbling}
Yichen Wang, Shangbin Feng, Abe~Bohan Hou, Xiao Pu, Chao Shen, Xiaoming Liu, Yulia Tsvetkov, and Tianxing He.
\newblock Stumbling blocks: Stress testing the robustness of machine-generated text detectors under attacks.
\newblock {\em arXiv preprint arXiv:2402.11638}, 2024.

\bibitem{wang2022ntk}
Yite Wang, Dawei Li, and Ruoyu Sun.
\newblock Ntk-sap: Improving neural network pruning by aligning training dynamics.
\newblock In {\em The Eleventh International Conference on Learning Representations}, 2022.

\bibitem{wang2023lemon}
Yite Wang, Jiahao Su, Hanlin Lu, Cong Xie, Tianyi Liu, Jianbo Yuan, Haibin Lin, Ruoyu Sun, and Hongxia Yang.
\newblock Lemon: Lossless model expansion.
\newblock In {\em The Twelfth International Conference on Learning Representations}, 2023.

\bibitem{wang2024balanced}
Yite Wang, Jing Wu, Naira Hovakimyan, and Ruoyu Sun.
\newblock Balanced training for sparse gans.
\newblock {\em Advances in Neural Information Processing Systems}, 36, 2024.

\bibitem{weaver2018regulation}
John~Frank Weaver.
\newblock Regulation of artificial intelligence in the united states.
\newblock In {\em Research Handbook on the Law of Artificial Intelligence}, pages 155--212. Edward Elgar Publishing, 2018.

\bibitem{webster2020measuring}
Kellie Webster, Xuezhi Wang, Ian Tenney, Alex Beutel, Emily Pitler, Ellie Pavlick, Jilin Chen, Ed~Chi, and Slav Petrov.
\newblock Measuring and reducing gendered correlations in pre-trained models.
\newblock {\em arXiv preprint arXiv:2010.06032}, 2020.

\bibitem{weidinger2021ethical}
Laura Weidinger, John Mellor, Maribeth Rauh, Conor Griffin, Jonathan Uesato, Po-Sen Huang, Myra Cheng, Mia Glaese, Borja Balle, Atoosa Kasirzadeh, et~al.
\newblock Ethical and social risks of harm from language models.
\newblock {\em arXiv preprint arXiv:2112.04359}, 2021.

\bibitem{weinberger2006graph}
Kilian~Q Weinberger, Fei Sha, Qihui Zhu, and Lawrence Saul.
\newblock Graph laplacian regularization for large-scale semidefinite programming.
\newblock {\em Advances in neural information processing systems}, 19, 2006.

\bibitem{welbl2021challenges}
Johannes Welbl, Amelia Glaese, Jonathan Uesato, Sumanth Dathathri, John Mellor, Lisa~Anne Hendricks, Kirsty Anderson, Pushmeet Kohli, Ben Coppin, and Po-Sen Huang.
\newblock Challenges in detoxifying language models.
\newblock {\em arXiv preprint arXiv:2109.07445}, 2021.

\bibitem{wen2023unveiling}
Jiaxin Wen, Pei Ke, Hao Sun, Zhexin Zhang, Chengfei Li, Jinfeng Bai, and Minlie Huang.
\newblock Unveiling the implicit toxicity in large language models.
\newblock {\em arXiv preprint arXiv:2311.17391}, 2023.

\bibitem{winograd2023loose}
Amy Winograd.
\newblock Loose-lipped large language models spill your secrets: The privacy implications of large language models.
\newblock {\em Harvard Journal of Law \& Technology}, 36(2), 2023.

\bibitem{wolf2023fundamental}
Yotam Wolf, Noam Wies, Yoav Levine, and Amnon Shashua.
\newblock Fundamental limitations of alignment in large language models.
\newblock {\em arXiv preprint arXiv:2304.11082}, 2023.

\bibitem{wu2023pairwise}
Tianhao Wu, Banghua Zhu, Ruoyu Zhang, Zhaojin Wen, Kannan Ramchandran, and Jiantao Jiao.
\newblock Pairwise proximal policy optimization: Harnessing relative feedback for llm alignment.
\newblock {\em arXiv preprint arXiv:2310.00212}, 2023.

\bibitem{wu2023brief}
Tianyu Wu, Shizhu He, Jingping Liu, Siqi Sun, Kang Liu, Qing-Long Han, and Yang Tang.
\newblock A brief overview of chatgpt: The history, status quo and potential future development.
\newblock {\em IEEE/CAA Journal of Automatica Sinica}, 10(5):1122--1136, 2023.

\bibitem{wu2023privacy}
Tong Wu, Ashwinee Panda, Jiachen~T Wang, and Prateek Mittal.
\newblock Privacy-preserving in-context learning for large language models.
\newblock {\em arXiv e-prints}, pages arXiv--2305, 2023.

\bibitem{xiang2021protecting}
Tao Xiang, Chunlong Xie, Shangwei Guo, Jiwei Li, and Tianwei Zhang.
\newblock Protecting your nlg models with semantic and robust watermarks.
\newblock {\em arXiv preprint arXiv:2112.05428}, 2021.

\bibitem{xiang2024badchain}
Zhen Xiang, Fengqing Jiang, Zidi Xiong, Bhaskar Ramasubramanian, Radha Poovendran, and Bo~Li.
\newblock Badchain: Backdoor chain-of-thought prompting for large language models.
\newblock In {\em ICLR}, 2024.

\bibitem{xiao2023large}
Yijia Xiao, Yiqiao Jin, Yushi Bai, Yue Wu, Xianjun Yang, Xiao Luo, Wenchao Yu, Xujiang Zhao, Yanchi Liu, Haifeng Chen, et~al.
\newblock Large language models can be good privacy protection learners.
\newblock {\em arXiv preprint arXiv:2310.02469}, 2023.

\bibitem{xu2021robust}
Han Xu, Xiaorui Liu, Yaxin Li, Anil Jain, and Jiliang Tang.
\newblock To be robust or to be fair: Towards fairness in adversarial training.
\newblock In {\em International conference on machine learning}, pages 11492--11501. PMLR, 2021.

\bibitem{xu2023training}
Mingbin Xu, Congzheng Song, Ye~Tian, Neha Agrawal, Filip Granqvist, Rogier van Dalen, Xiao Zhang, Arturo Argueta, Shiyi Han, Yaqiao Deng, et~al.
\newblock Training large-vocabulary neural language models by private federated learning for resource-constrained devices.
\newblock In {\em ICASSP 2023-2023 IEEE International Conference on Acoustics, Speech and Signal Processing (ICASSP)}, pages 1--5. IEEE, 2023.

\bibitem{yan2023comprehensive}
Hao Yan, Chaozhuo Li, Ruosong Long, Chao Yan, Jianan Zhao, Wenwen Zhuang, Jun Yin, Peiyan Zhang, Weihao Han, Hao Sun, et~al.
\newblock A comprehensive study on text-attributed graphs: Benchmarking and rethinking.
\newblock {\em NeurIPS}, 36:17238--17264, 2023.

\bibitem{yan2020interpretable}
Li~Yan, Hai-Tao Zhang, Jorge Goncalves, Yang Xiao, Maolin Wang, Yuqi Guo, Chuan Sun, Xiuchuan Tang, Liang Jing, Mingyang Zhang, et~al.
\newblock An interpretable mortality prediction model for covid-19 patients.
\newblock {\em Nature machine intelligence}, 2(5):283--288, 2020.

\bibitem{yang2024editworld}
Ling Yang, Bohan Zeng, Jiaming Liu, Hong Li, Minghao Xu, Wentao Zhang, and Shuicheng Yan.
\newblock Editworld: Simulating world dynamics for instruction-following image editing.
\newblock {\em arXiv:2405.14785}, 2024.

\bibitem{yang2023alignment}
Yuqing Yang, Ethan Chern, Xipeng Qiu, Graham Neubig, and Pengfei Liu.
\newblock Alignment for honesty.
\newblock {\em arXiv preprint arXiv:2312.07000}, 2023.

\bibitem{yang2022unified}
Zonghan Yang, Xiaoyuan Yi, Peng Li, Yang Liu, and Xing Xie.
\newblock Unified detoxifying and debiasing in language generation via inference-time adaptive optimization.
\newblock {\em arXiv preprint arXiv:2210.04492}, 2022.

\bibitem{yao2023promptcare}
Hongwei Yao, Jian Lou, Kui Ren, and Zhan Qin.
\newblock Promptcare: Prompt copyright protection by watermark injection and verification.
\newblock {\em arXiv preprint arXiv:2308.02816}, 2023.

\bibitem{yao2023large}
Yuanshun Yao, Xiaojun Xu, and Yang Liu.
\newblock Large language model unlearning.
\newblock {\em arXiv preprint arXiv:2310.10683}, 2023.

\bibitem{yao2023editing}
Yunzhi Yao, Peng Wang, Bozhong Tian, Siyuan Cheng, Zhoubo Li, Shumin Deng, Huajun Chen, and Ningyu Zhang.
\newblock Editing large language models: Problems, methods, and opportunities.
\newblock {\em arXiv preprint arXiv:2305.13172}, 2023.

\bibitem{ye2023multiplexed}
Jiachi Ye, Haoyan Kang, Hao Wang, Salem Altaleb, Elham Heidari, Navid Asadizanjani, Volker~J Sorger, and Hamed Dalir.
\newblock Multiplexed oam beams classification via fourier optical convolutional neural network.
\newblock In {\em 2023 IEEE Photonics Conference (IPC)}, pages 1--2. IEEE, 2023.

\bibitem{ye2023oam}
Jiachi Ye, Haoyan Kang, Hao Wang, Salem Altaleb, Elham Heidari, Navid Asadizanjani, Volker~J Sorger, and Hamed Dalir.
\newblock Oam beams multiplexing and classification under atmospheric turbulence via fourier convolutional neural network.
\newblock In {\em Frontiers in Optics}, pages JTu4A--73. Optica Publishing Group, 2023.

\bibitem{ye2023demultiplexing}
Jiachi Ye, Haoyan Kang, Hao Wang, Chen Shen, Belal Jahannia, Elham Heidari, Navid Asadizanjani, Mohammad-Ali Miri, Volker~J Sorger, and Hamed Dalir.
\newblock Demultiplexing oam beams via fourier optical convolutional neural network.
\newblock In {\em Laser Beam Shaping XXIII}, volume 12667, pages 16--33. SPIE, 2023.

\bibitem{ye2023free}
Jiachi Ye, Maria Solyanik, Zibo Hu, Hamed Dalir, Behrouz~Movahhed Nouri, and Volker~J Sorger.
\newblock Free-space optical multiplexed orbital angular momentum beam identification system using fourier optical convolutional layer based on 4f system.
\newblock In {\em Complex Light and Optical Forces XVII}, volume 12436, pages 70--80. SPIE, 2023.

\bibitem{Yoo2023RobustMN}
Kiyoon Yoo, Wonhyuk Ahn, Jiho Jang, and No~Jun Kwak.
\newblock Robust multi-bit natural language watermarking through invariant features.
\newblock In {\em Annual Meeting of the Association for Computational Linguistics}, 2023.

\bibitem{yu2023unlearning}
Charles Yu, Sullam Jeoung, Anish Kasi, Pengfei Yu, and Heng Ji.
\newblock Unlearning bias in language models by partitioning gradients.
\newblock In {\em Findings of the Association for Computational Linguistics: ACL 2023}, pages 6032--6048, 2023.

\bibitem{yu2021differentially}
Da~Yu, Saurabh Naik, Arturs Backurs, Sivakanth Gopi, Huseyin~A Inan, Gautam Kamath, Janardhan Kulkarni, Yin~Tat Lee, Andre Manoel, Lukas Wutschitz, et~al.
\newblock Differentially private fine-tuning of language models.
\newblock {\em arXiv preprint arXiv:2110.06500}, 2021.

\bibitem{yu2023mixup}
Liu Yu, Yuzhou Mao, Jin Wu, and Fan Zhou.
\newblock Mixup-based unified framework to overcome gender bias resurgence.
\newblock In {\em Proceedings of the 46th International ACM SIGIR Conference on Research and Development in Information Retrieval}, SIGIR '23, page 1755–1759, New York, NY, USA, 2023. Association for Computing Machinery.

\bibitem{yu2023bag}
Weichen Yu, Tianyu Pang, Qian Liu, Chao Du, Bingyi Kang, Yan Huang, Min Lin, and Shuicheng Yan.
\newblock Bag of tricks for training data extraction from language models.
\newblock {\em arXiv preprint arXiv:2302.04460}, 2023.

\bibitem{zayed2023should}
Abdelrahman Zayed, Goncalo Mordido, Samira Shabanian, and Sarath Chandar.
\newblock Should we attend more or less? modulating attention for fairness.
\newblock {\em arXiv preprint arXiv:2305.13088}, 2023.

\bibitem{zayed2023deep}
Abdelrahman Zayed, Prasanna Parthasarathi, Gon{\c{c}}alo Mordido, Hamid Palangi, Samira Shabanian, and Sarath Chandar.
\newblock Deep learning on a healthy data diet: Finding important examples for fairness.
\newblock In {\em Proceedings of the AAAI Conference on Artificial Intelligence}, volume~37, pages 14593--14601, 2023.

\bibitem{zeng2022glm}
Aohan Zeng, Xiao Liu, Zhengxiao Du, Zihan Wang, Hanyu Lai, Ming Ding, Zhuoyi Yang, Yifan Xu, Wendi Zheng, Xiao Xia, et~al.
\newblock Glm-130b: An open bilingual pre-trained model.
\newblock {\em arXiv preprint arXiv:2210.02414}, 2022.

\bibitem{zeng2022mpcvit}
Wenxuan Zeng, Meng Li, Wenjie Xiong, Wenjie Lu, Jin Tan, Runsheng Wang, and Ru~Huang.
\newblock Mpcvit: Searching for mpc-friendly vision transformer with heterogeneous attention.
\newblock {\em arXiv preprint arXiv:2211.13955}, 2022.

\bibitem{zhang2021survey}
Chen Zhang, Yu~Xie, Hang Bai, Bin Yu, Weihong Li, and Yuan Gao.
\newblock A survey on federated learning.
\newblock {\em Knowledge-Based Systems}, 216:106775, 2021.

\bibitem{zhang2022augmented}
Chi Zhang, Sotthiwat Ekanut, Liangli Zhen, and Zengxiang Li.
\newblock Augmented multi-party computation against gradient leakage in federated learning.
\newblock {\em IEEE Transactions on Big Data}, 2022.

\bibitem{zhang2023multi}
Meiying Zhang, Huan Zhao, Sheldon Ebron, Ruitao Xie, and Kan Yang.
\newblock Multi-criteria client selection and scheduling with fairness guarantee for federated learning service.
\newblock {\em arXiv preprint arXiv:2312.14941}, 2023.

\bibitem{zhang2024high}
Peiyan Zhang, Chaozhuo Li, Liying Kang, Feiran Huang, Senzhang Wang, Xing Xie, and Sunghun Kim.
\newblock High-frequency-aware hierarchical contrastive selective coding for representation learning on text-attributed graphs.
\newblock {\em arXiv:2402.16240}, 2024.

\bibitem{zhang2023foundation}
Peiyan Zhang, Haoyang Liu, Chaozhuo Li, Xing Xie, Sunghun Kim, and Haohan Wang.
\newblock Foundation model-oriented robustness: Robust image model evaluation with pretrained models.
\newblock In {\em ICLR}, 2023.

\bibitem{zhang2022opt}
Susan Zhang, Stephen Roller, Naman Goyal, Mikel Artetxe, Moya Chen, Shuohui Chen, Christopher Dewan, Mona Diab, Xian Li, Xi~Victoria Lin, et~al.
\newblock Opt: Open pre-trained transformer language models.
\newblock {\em arXiv preprint arXiv:2205.01068}, 2022.

\bibitem{zhang2024unlocking}
Ye~Zhang, Kailin Gui, Mengran Zhu, Yong Hao, and Haozhan Sun.
\newblock Unlocking personalized anime recommendations: Langchain and llm at the forefront.
\newblock {\em Journal of Industrial Engineering and Applied Science}, 2(2):46--53, 2024.

\bibitem{zhang2019dialogpt}
Yizhe Zhang, Siqi Sun, Michel Galley, Yen-Chun Chen, Chris Brockett, Xiang Gao, Jianfeng Gao, Jingjing Liu, and Bill Dolan.
\newblock Dialogpt: Large-scale generative pre-training for conversational response generation.
\newblock {\em arXiv preprint arXiv:1911.00536}, 2019.

\bibitem{zhang2023siren}
Yue Zhang, Yafu Li, Leyang Cui, Deng Cai, Lemao Liu, Tingchen Fu, Xinting Huang, Enbo Zhao, Yu~Zhang, Yulong Chen, et~al.
\newblock Siren's song in the ai ocean: A survey on hallucination in large language models.
\newblock {\em arXiv:2309.01219}, 2023.

\bibitem{zhang2023c2pi}
Yuke Zhang, Dake Chen, Souvik Kundu, Haomei Liu, Ruiheng Peng, and Peter~A Beerel.
\newblock C2pi: An efficient crypto-clear two-party neural network private inference.
\newblock {\em arXiv preprint arXiv:2304.13266}, 2023.

\bibitem{zhang2023ethicist}
Zhexin Zhang, Jiaxin Wen, and Minlie Huang.
\newblock Ethicist: Targeted training data extraction through loss smoothed soft prompting and calibrated confidence estimation.
\newblock {\em arXiv preprint arXiv:2307.04401}, 2023.

\bibitem{zhang2023fedpetuning}
Zhuo Zhang, Yuanhang Yang, Yong Dai, Qifan Wang, Yue Yu, Lizhen Qu, and Zenglin Xu.
\newblock Fedpetuning: When federated learning meets the parameter-efficient tuning methods of pre-trained language models.
\newblock In {\em Annual Meeting of the Association of Computational Linguistics 2023}, pages 9963--9977. Association for Computational Linguistics (ACL), 2023.

\bibitem{zhao2023explainability}
Haiyan Zhao, Hanjie Chen, Fan Yang, Ninghao Liu, Huiqi Deng, Hengyi Cai, Shuaiqiang Wang, Dawei Yin, and Mengnan Du.
\newblock Explainability for large language models: A survey.
\newblock {\em ACM Transactions on Intelligent Systems and Technology}, 2023.

\bibitem{zhao2023competeai}
Qinlin Zhao, Jindong Wang, Yixuan Zhang, Yiqiao Jin, Kaijie Zhu, Hao Chen, and Xing Xie.
\newblock Competeai: Understanding the competition behaviors in large language model-based agents.
\newblock {\em arXiv preprint arXiv:2310.17512}, 2023.

\bibitem{zheng2023judging}
Lianmin Zheng, Wei-Lin Chiang, Ying Sheng, Siyuan Zhuang, Zhanghao Wu, Yonghao Zhuang, Zi~Lin, Zhuohan Li, Dacheng Li, Eric Xing, et~al.
\newblock Judging llm-as-a-judge with mt-bench and chatbot arena.
\newblock {\em arXiv preprint arXiv:2306.05685}, 2023.

\bibitem{zheng2023primer}
Mengxin Zheng, Qian Lou, and Lei Jiang.
\newblock Primer: Fast private transformer inference on encrypted data.
\newblock {\em arXiv preprint arXiv:2303.13679}, 2023.

\bibitem{zheng2023does}
Shen Zheng, Jie Huang, and Kevin Chen-Chuan Chang.
\newblock Why does chatgpt fall short in answering questions faithfully?
\newblock {\em arXiv preprint arXiv:2304.10513}, 2023.

\bibitem{zhou2023lima}
Chunting Zhou, Pengfei Liu, Puxin Xu, Srini Iyer, Jiao Sun, Yuning Mao, Xuezhe Ma, Avia Efrat, Ping Yu, Lili Yu, et~al.
\newblock Lima: Less is more for alignment.
\newblock {\em arXiv preprint arXiv:2305.11206}, 2023.

\bibitem{zhou2023navigating}
Kaitlyn Zhou, Dan Jurafsky, and Tatsunori Hashimoto.
\newblock Navigating the grey area: Expressions of overconfidence and uncertainty in language models.
\newblock {\em arXiv preprint arXiv:2302.13439}, 2023.

\bibitem{zhu2023promptbench}
Kaijie Zhu, Jindong Wang, Jiaheng Zhou, Zichen Wang, Hao Chen, Yidong Wang, Linyi Yang, Wei Ye, Neil~Zhenqiang Gong, Yue Zhang, et~al.
\newblock Promptbench: Towards evaluating the robustness of large language models on adversarial prompts.
\newblock {\em arXiv preprint arXiv:2306.04528}, 2023.

\bibitem{wnebo2022optimizing}
Wenbo Zhu.
\newblock {Optimizing distributed networking with big data scheduling and cloud computing}.
\newblock In Warwick Powell and Amr Tolba, editors, {\em International Conference on Cloud Computing, Internet of Things, and Computer Applications (CICA 2022)}, volume 12303, page 1230306. International Society for Optics and Photonics, SPIE, 2022.

\bibitem{wenbozhu2021}
Wenbo Zhu and Tiechuan Hu.
\newblock Twitter sentiment analysis of covid vaccines.
\newblock In {\em 2021 5th International Conference on Artificial Intelligence and Virtual Reality (AIVR)}, AIVR 2021, page 118–122, New York, NY, USA, 2021. Association for Computing Machinery.

\bibitem{zhuang2024robust}
Jun Zhuang.
\newblock Robust data-centric graph structure learning for text classification.
\newblock In {\em Companion Proceedings of the ACM on Web Conference 2024}, pages 1486--1495, 2024.

\bibitem{zhuang2024understanding}
Jun Zhuang and Casey Kennington.
\newblock Understanding survey paper taxonomy about large language models via graph representation learning.
\newblock {\em arXiv preprint arXiv:2402.10409}, 2024.

\bibitem{zou2023universal}
Andy Zou, Zifan Wang, J~Zico Kolter, and Matt Fredrikson.
\newblock Universal and transferable adversarial attacks on aligned language models.
\newblock {\em arXiv preprint arXiv:2307.15043}, 2023.

\bibitem{zou2023jointmatch}
Henry Zou and Cornelia Caragea.
\newblock Jointmatch: A unified approach for diverse and collaborative pseudo-labeling to semi-supervised text classification.
\newblock In {\em Proceedings of the 2023 Conference on Empirical Methods in Natural Language Processing}, pages 7290--7301, 2023.

\bibitem{zou2024eiven}
Henry~Peng Zou, Gavin~Heqing Yu, Ziwei Fan, Dan Bu, Han Liu, Peng Dai, Dongmei Jia, and Cornelia Caragea.
\newblock Eiven: Efficient implicit attribute value extraction using multimodal llm.
\newblock In {\em Proceedings of the 2024 Conference of the North American Chapter of the Association for Computational Linguistics: Human Language Technologies: Industry Track}, 2024.

\bibitem{zou2023decrisismb}
Henry~Peng Zou, Yue Zhou, Weizhi Zhang, and Cornelia Caragea.
\newblock Decrisismb: Debiased semi-supervised learning for crisis tweet classification via memory bank.
\newblock In {\em The 2023 Conference on Empirical Methods in Natural Language Processing}, 2023.

\end{thebibliography}

\end{document}